\documentclass[journal]{IEEEtran}
\usepackage{cite}
\ifCLASSINFOpdf
   \usepackage[pdftex]{graphicx}
   \graphicspath{{img/}}
   \newcommand{\bcaption}{%
\setlength{\abovecaptionskip}{0pt}%
\setlength{\belowcaptionskip}{2pt}%
\caption}

\else
\fi
\usepackage{amsmath}
\usepackage{amssymb}
\usepackage{multirow}
\usepackage{color}
\usepackage{diagbox}
\usepackage{algorithmic}
\usepackage{array}
\ifCLASSOPTIONcompsoc
 \usepackage[caption=false,font=normalsize,labelfont=sf,textfont=sf]{subfig}
\else
 \usepackage[caption=false,font=footnotesize]{subfig}
\fi

\ifCLASSOPTIONcaptionsoff
 \usepackage[nomarkers]{endfloat}
\let\MYoriglatexcaption\caption
\renewcommand{\caption}[2][\relax]{\MYoriglatexcaption[#2]{#2}}
\fi
\begin{document}
\title{One-shot Key Information Extraction from Document with Deep Partial Graph Matching}

\DeclareRobustCommand*{\IEEEauthorrefmarkN}[1]{%
  \raisebox{0pt}[0pt][0pt]{\textsuperscript{\footnotesize #1}}%
}

\author{Minghong~Yao\IEEEauthorrefmarkN{1}\IEEEauthorrefmark{1},
    Zhiguang~Liu\IEEEauthorrefmarkN{2}\IEEEauthorrefmark{1},
    Liangwei~Wang\IEEEauthorrefmarkN{2},
    Houqiang~Li\IEEEauthorrefmarkN{1}
    and~Liansheng~Zhuang\IEEEauthorrefmarkN{1}\IEEEauthorrefmark{2}
        % ~\IEEEmembership{Life~Fellow,~IEEE}% <-this % stops a space
        
\IEEEauthorblockA{University of Science and Technology of China\IEEEauthorrefmarkN{1} and Noah's Ark Lab, Huawei Technologies, China\IEEEauthorrefmarkN{2}\\
mhyao1@mail.ustc.edu.cn,
\{liuzhiguang1,wangliangwei\}@huawei.com,
},
\{lihq,lszhuang\}@ustc.edu.cn
\thanks{\IEEEauthorrefmark{1}Equal contribution.}
\thanks{\IEEEauthorrefmark{2}Corresponding author.}}

\maketitle

\begin{abstract}
Automating the Key Information Extraction (KIE) from documents improves efficiency, productivity, and security in many industrial scenarios such as rapid indexing and archiving. Many existing supervised learning methods for the KIE task need to feed a large number of labeled samples and learn separate models for different types of documents. However, collecting and labeling a large dataset is time-consuming and is not a user-friendly requirement for many cloud platforms. To overcome these challenges, we propose a deep end-to-end trainable network for one-shot KIE using partial graph matching. Contrary to previous methods that the learning of similarity and solving are optimized separately, our method enables the learning of the two processes in an end-to-end framework. Existing one-shot KIE methods are either template or simple attention-based learning approach that struggle to handle texts that are shifted beyond their desired positions caused by printers, as illustrated in Fig.~\ref{fig:spatial_drift}. To solve this problem, we add one-to-(at most)-one constraint such that we will find the globally optimized solution even if some texts are drifted.  Further, we design a multimodal context ensemble block to boost the performance through fusing features of spatial, textual, and aspect representations. To promote research of KIE, we collected and annotated a one-shot document KIE dataset named DKIE with diverse types of images. The DKIE dataset consists of 2.5K document images captured by mobile phones in natural scenes, and it is the largest available one-shot KIE dataset up to now. The results of experiments on DKIE show that our method achieved state-of-the-art performance compared with recent one-shot and supervised learning approaches. The dataset and proposed one-shot KIE model will be released soon.
\end{abstract}


% \fi
\IEEEpeerreviewmaketitle

% 标红的内容为在MM初稿的基础上新加的
\section{Introduction}
Companies often face the problem of searching through \textcolor{black}{and extracting information that they are interested in,} from their unorganized mix of physical paper and digital documents. This process can be time-consuming and tedious. \textcolor{black}{To automate this process, people studied the Key Information Extraction (KIE) task~\cite{cheng2020one,rusinol2013field,2019One}.} Thus, KIE is crucial to a company in terms of efficiency and productivity, and it has been successfully used in many industrial scenarios, such as fast indexing and efficient archiving.

% 这段话被我舍弃了
% Digitizing all the paper documents makes the first step to feeding the data in these documents into Key Information Extraction (KIE) algorithms. 
% KIE is the process of extracting structured information from document images.

\begin{figure*}[t]
	\centering
	\includegraphics[width = 0.9\linewidth]{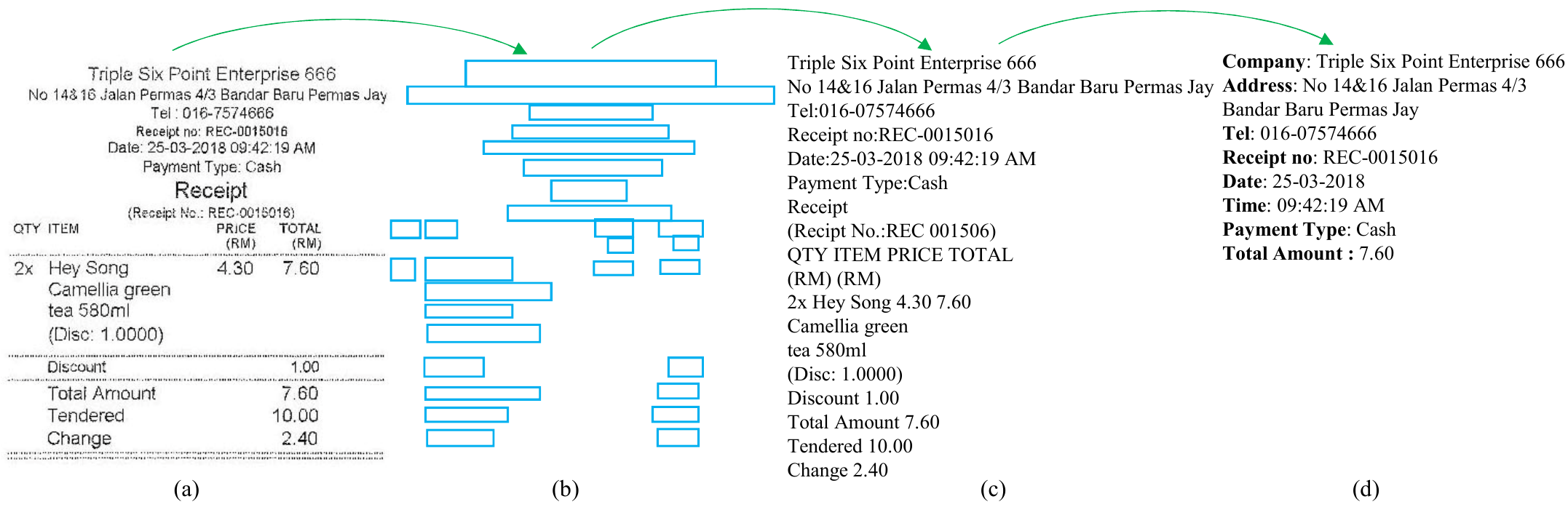}\\
	\bcaption{\label{fig:field-labeling-pipline}The pip-line of extracting user predefined key information. (a) Receipt, (b) Text detection, (c) Text recognition, (d) Text field labeling. }
\end{figure*}

A typical KIE consists of three key steps: text detection, text recognition, and \textcolor{black}{text field labeling}, as shown in Fig.~\ref{fig:field-labeling-pipline}. While the text detection and recognition approaches~\cite{yang2018inceptext,wan2020textscanner,yue2020robustscanner} have been studied widely in the area of Optical Character Recognition (OCR), \textcolor{black}{one-shot learning based text field labeling} is less studied. \textcolor{black}{The text field labeling task aims to identify the predefined label of each text field.} 

\begin{figure}[t]
	\centering
	\includegraphics[width = 0.7\linewidth]{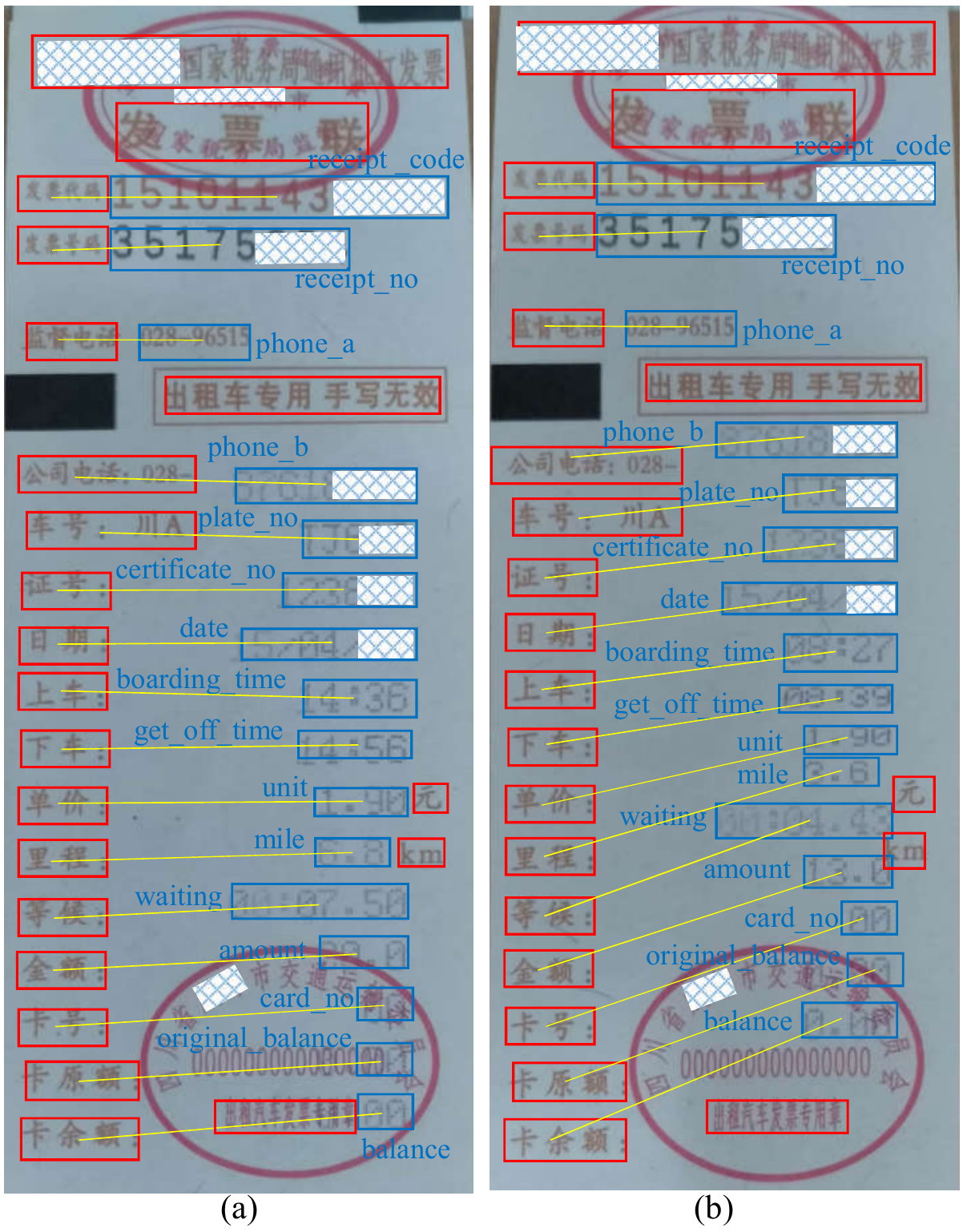}\\
	\bcaption{\label{fig:spatial_drift}\textcolor{black}{Samples containing drifted fields in the d0 dataset.} (a) Support document. (b) Query document. The red boxes represent landmarks (static zones), and the blue ones indicate fields (dynamic zones). \textcolor{black}{The users annotate support fields in the support document with predefined labels. Then, one can infer the labels of query fields in query documents with three steps. First, one can align query fields with their corresponding landmarks as the yellow lines did. Then, one can find support fields that align to the same landmarks, and therefore a mapping between a support field and a query field exists if they align to the same landmark. At last, one can use the labels of support fields as the labels of their mapped query fields.} The drifted fields in a document are likely to misalign with their corresponding landmarks caused by a printer.}
\end{figure}
% 我们的目标是query蓝框的类别，根据support中的
	% query中对齐landmark，用文字匹配  
	% 通过计算query中field与support中field哪个最相似，从而决定query中field的标签。

The layout of a document plays a key role in distinguishing different fields. Generally, as illustrated in Fig.~\ref{fig:field-labeling-pipline}, the \textit{Total}\_\textit{Amount} \textit{7.60} is much more likely on the above of \textit{Tendered}\ \textit{10.00}. Fig.~\ref{fig:our_dataset} shows some document images with different layouts and categories. Many learning-based methods~\cite{d2018field,xu2020layoutlm,Yu2020PICKPK,vrdgcn_naacl19,qian-etal-2019-graphie} have been proposed to utilize both the text and visual patterns for the KIE task. \textcolor{black}{They have shown good performance, but they require sufficient training data.} To reduce the cost of labor and alleviate the dependence of a large amount of training data for each type of document with a separate model, one-shot learning methods are studied. Early attempts at one-shot methods~\cite{chiticariu2013rule,schuster2013intellix,rusinol2013field,d2018field} are usually based on template for entity extraction. However, these rule-based methods are limited to specific layouts and are not general enough to scale to all types of documents. Cheng et al.\cite{cheng2020one} proposed an attention-based learning approach to transfer the spatial relationships between landmarks and fields from a support document to a query document. \textcolor{black}{However, their method cannot deal with drifted fields and outliers. In practice, printed docs often contain drifted fields and outliers as shown in Fig.~\ref{fig:spatial_drift}. Drifted fields refer to fields that are printed in unexpected positions. Thus the spatial relationships between landmarks and drifted fields are different from the one between landmarks and non-drifted fields. A direct transfer from the support documents to the query documents would fail because of this difference. Outliers refer to fields that do not match any fields in the support doc such as unexpected handwritten words. Their method cannot pick out the outliers too.} 

\textcolor{black}{To address the challenges of drifted fields and outliers, we propose to cast the text field labeling task as a partial graph matching problem. Our method uses multiple features such as position, shape, and text embedding, to measure the similarity score between a support and a query field. Then our method maps support fields to query fields to maximize the summation of similarity score of all mapped pairs of fields. Particularly, our method will obey the one-to-(at most)-one mapping constraint when it searches for the mapping between fields. This constraint can help map drifted query fields to the correct support fields even if there are other more similar support fields. Our method maps the outliers to no fields too.} 

The major contributions of this paper can be summarized as follows:
\begin{itemize}
    % \item We designed a spatial-shift robust attention module to learn more reasonable affinity for text field pairs between support and query documents.
    % \item By adapting Black Box Graph Matching (BB-GM) framework and Zero-Assignment Constraint Graph Matching (ZAC-GM) solver, we can cast Text Field Labeling (TFL) task into ZAC-GM task. One of the advantages of casting TFL task into ZAC-GM task is that our model can handle more difficult documents. What's more, we can also use the highly optimized GM solver to find the global solutions for TFL task and this can lead to stronger performance in practise. 
    
    \item We propose a deep end-to-end trainable network for one-shot Key Information Extraction (KIE) using \textcolor{black}{partial} graph matching with the one-to-(at most)-one mapping constraint. Our method enables the learning of similarity and solving for combinatorial optimization done in an end-to-end framework instead of solving these two phases explicitly separated as opposed to many previous methods. To the best of our knowledge, this is the first KIE approach that generates globally optimized solutions.
    
    \item We design a simple context ensemble block to fuse features of spatial, textual, and aspect representations. The proposed framework is general enough to plugin other constraints such as zero assignment constraint to adapt to different KIE tasks.
    
    \item To promote research in KIE, one dataset is constructed and the proposed one-shot KIE model will be released soon. Note that these datasets cover diverse types of document images, and much of them are highly difficult with spatial drift.
    
    \item Our method achieves state-of-the-art performance on the collected datasets. 
    
    % With the use of the highly optimized GM solver, our method process a typical document image with 12 fields within 200 ms.
\end{itemize}

\begin{figure*}[t]
	\centering
	\includegraphics[width = 0.9\linewidth]{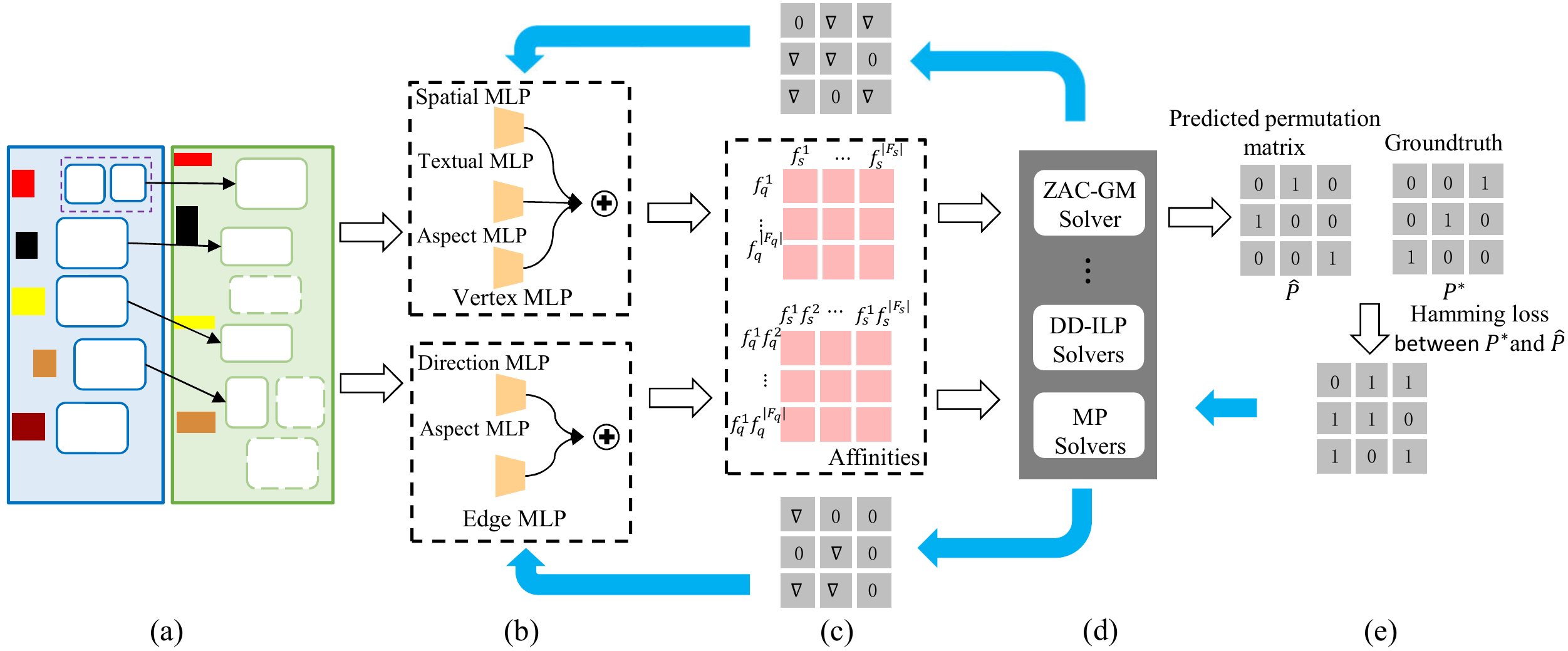}\\
	\bcaption{\label{fig:system} Overview of the proposed model. In step (a), we build the graphs, extract and concatenate vertex and edge features. In step (b), we feed different features into separate Multi-layer Perceptrons (MLP), and their outputs are vertex and edge affinity matrices. In step (c), we compute the average vertex and edge affinity matrices over different MLPs. In step (d), the average vertex and edge affinity matrices are feed into the combinatorial solvers such as ZAC-GM solvers, and its output is $\hat{P}$, which is called the predicted permutation matrix. The elements of $\hat{P}$ are 1 or 0. In step (e), we calculate the hamming loss between $\hat{P}$ and the groundtruth $P^*$. Lastly, we compute the gradients of hamming loss for the parameters of MLPs. Each $\nabla$ in the gradient matrix means the corresponding element is non-zero.}
\end{figure*}

\section{Related Work}
In this section, we first review previous work on keyword spotting task, KIE, one-shot learning of KIE, and then discuss approaches for graph matching that inspired our approach. 

% 讨论其他的Document analysis方法
% \subsection{Document analysis}
% \textcolor{black}{Many existing KIE approaches treat this labeling task as a pure Named Entity Recognition (NER) task using the sequence tagging methods~\cite{lample2016neural}. The NER methods~\cite{strakova-etal-2019-neural,jiao-etal-2020-tinybert} work on plain text and ignore other important information such as document layout and visual cues.  However, the task of KIE is more challenging due to the complexity of various inputs such as text, layout, and visual cues.}

\subsection{Keyword Spotting}
\textcolor{black}{The methods in KeyWord Spotting (KWS)~\cite{vidal2021probabilistic} task cannot solve the KIE task. On the one hand, KWS checks if a given text exists in an image and finds its location. On the other hand, KIE aims at assigning a label to each field based on text detection and recognition results, e.g., identifying “Tom” as “name” in an ID card. However, KWS can’t find “Tom” in the support doc because the name can vary for different ID cards. Thus, both the methods and datasets for KWS are not suitable for KIE.}

\subsection{Key Information Extraction}
Language model based methods~\cite{devlin2018bert,dai-etal-2019-transformer,yang2019xlnet} work on plain text representations. However, document layout information is also crucial for information extraction. Then, many existing learning-based methods \cite{denk2019bertgrid,katti-etal-2018-chargrid,palm2019attend} tend to use both textual and visual embedding to enhance the performance of KIE.

% \textcolor{black}{Zhao et al.\cite{d2018field} applied convolution operations on the gridded texts that preserve texts' relative spatial relationship in a document.} \textcolor{black}{Qian et al.\cite{qian-etal-2019-graphie} introduced GraphIE to explore a broad set of dependencies between words using graph convolutions.} 

Liu et al.\cite{vrdgcn_naacl19} introduced a method that combines visual and textual information in an image by a graph convolution model. Yu et al.\cite{Yu2020PICKPK} presented a layout extraction framework via combining graph learning with graph convolutions, which resulted in rich semantic representations of textual, visual, and layout representations. Zhang et al.\cite{zhang2020trie} fused the embedding of visual and textual representations such that the two tasks can reinforce each learning process. Inspired by BERT\cite{devlin2018bert}, Xu et al.\cite{xu2020layoutlm} proposed a pre-training method that jointly models the text and layout information within a single framework. However, this method requires explicit segmentation of individual words such that some modern OCR approaches are not applicable.

While the approaches we discussed above achieved promising results, we have to train a separate model for each type of document that is a waste of resources. Additionally, we have to collect and manually annotate a large number of labeled images for each category of document, which is labor-intensive and time-consuming.

\subsection{One-shot Learning of KIE}

Medvet et al.\cite{medvet2011probabilistic} proposed a probabilistic model to search key information from a document. However, their method required two sequences have the same length. Rusinol et al.\cite{rusinol2013field} presented an iterative framework to extract information from administrative documents. They introduced a star graph to model the spatial relationships among different fields. The weights for each node were adapted by term frequency-inverse document frequency (TF-IDF). However, for some scenarios such as invoice, where most words are digits such that TF-IDF is not robust enough.

Cheng et al.\cite{cheng2020one} presented a one-shot field labeling method using attention and belief propagation to retrieve structured information. Although their method dramatically simplifies the labeling process and achieved good performance compared with previous one-shot-based approaches, the final matching results are not globally optimal. For example, as illustrated in Fig.~\ref{fig:spatial_drift}, $phone\_b$ and $plate\_no$ were labeled as the same class due to the vertical drift caused by the printer.

Existing one-shot approaches are mostly rule-based and struggled to identify text fields close to each other. In particular, the performance of crucial information extraction dropped sharply when large spatial drift is observed between the landmark and corresponding fields.  These performance drops suggesting that exiting models are sensitive to spatial relationship variations. This paper proposes a deep end-to-end trainable structured information extraction framework that is topology invariant and global optimized such that cases like two different fields are mapped to the same category would be alleviated.

\subsection{Graph Matching}
Graph matching approaches have been widely used in computer vision tasks, such as key-points matching. In this subsection, we focus on deep learning methods for graph matching.

Hammami et al.\cite{hammami2015one} proposed a subgraph isomorphism-based method to extract informative areas in administrative and commercial forms using color information. The information extraction task is then converted to search the sub-graph of a query document for the best matching of the graph representation of the supporting document. However, many documents are scanned in black and white that limits the application of this method.

Andrei Zanfir and Cristian Sminchisescu\cite{zanfir2018deep} proposed to model deep feature extraction and solve combinatorial optimization as an end-to-end learning framework. Wang et al.\cite{wang2019learning} presented an end-to-end differentiable deep combinatorial learning of graph matching. Different from the pixel offset loss\cite{zanfir2018deep}, a permutation loss based on Sinkhorn net was employed to handle an arbitrary number of nodes for combinatorial graph matching. Further, Wang et al.\cite{wang2020learning} embedded the learning of affinities and into a uniform framework instead of solving them separately\cite{zanfir2018deep}.

\begin{figure*}[!t]
	\centering
	\subfloat[Samples contain multi-region fields.]{\includegraphics[width=0.3151\linewidth]{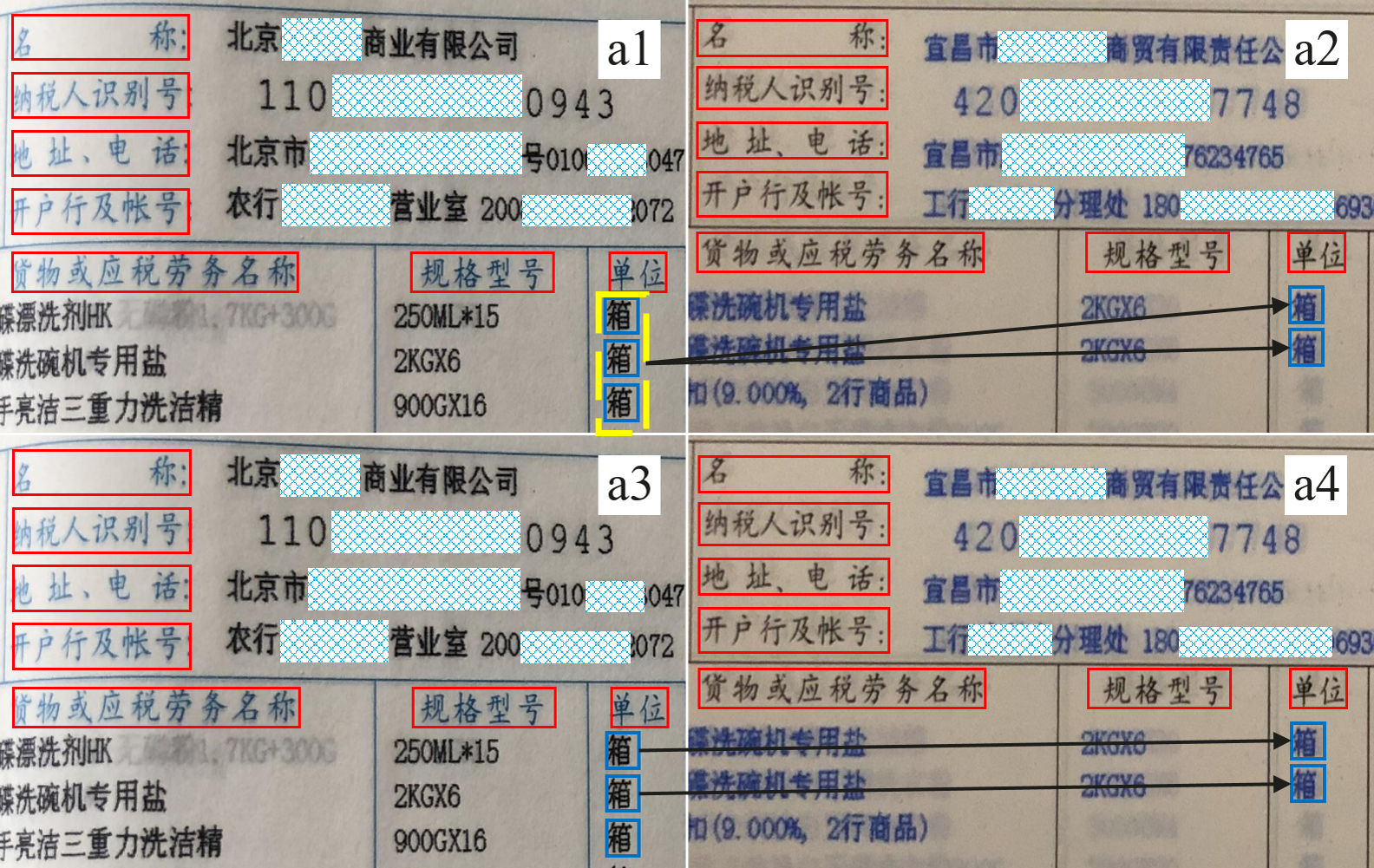}%
		\label{fig_first_case}}
% 	\hfil
	\subfloat[Samples contain drifted fields.]{\includegraphics[width=0.368\linewidth]{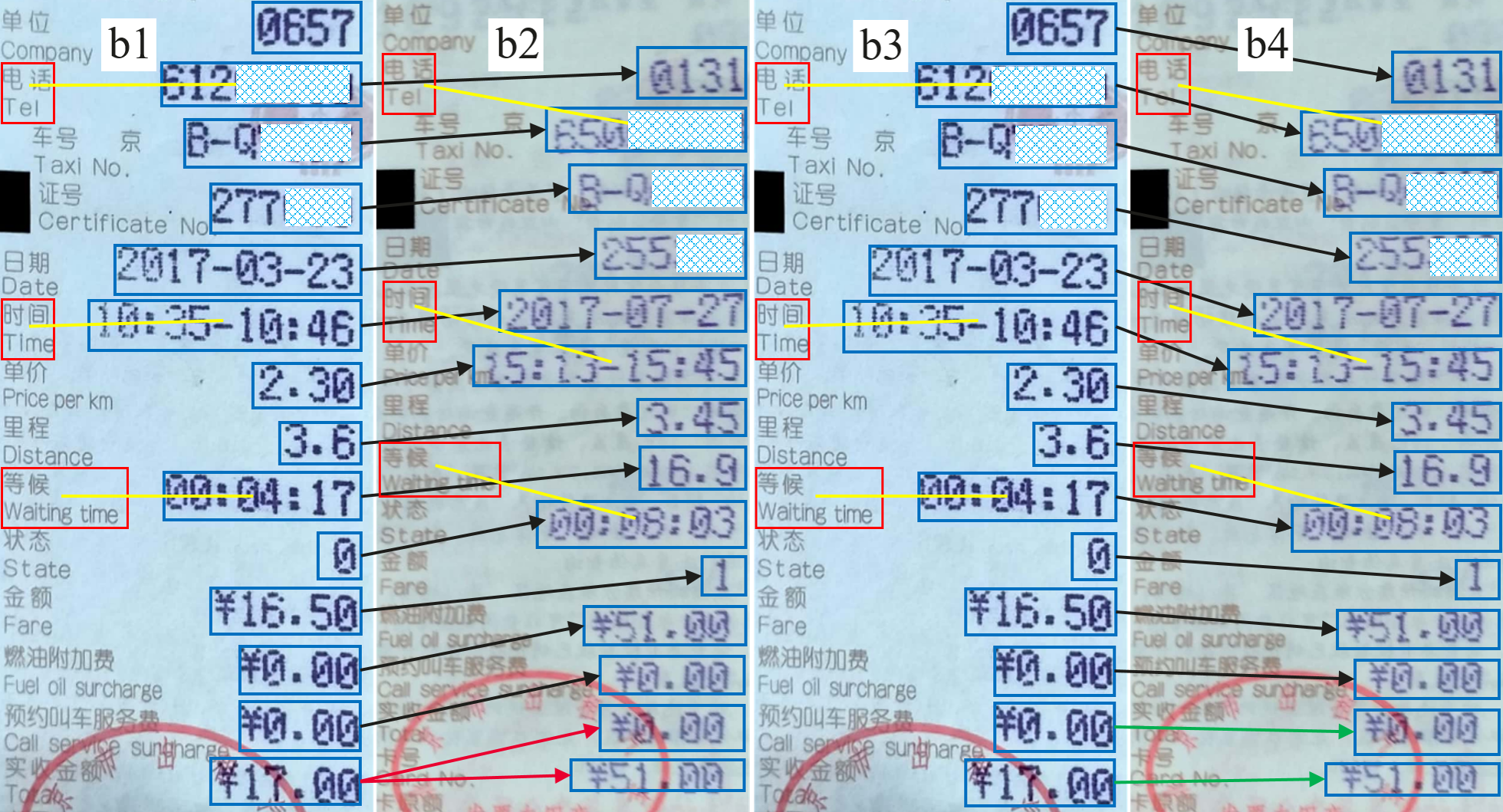}%
		\label{fig_second_case}}
% 	\hfil
	\subfloat[Samples contain outliers.]{\includegraphics[width=0.3151\linewidth]{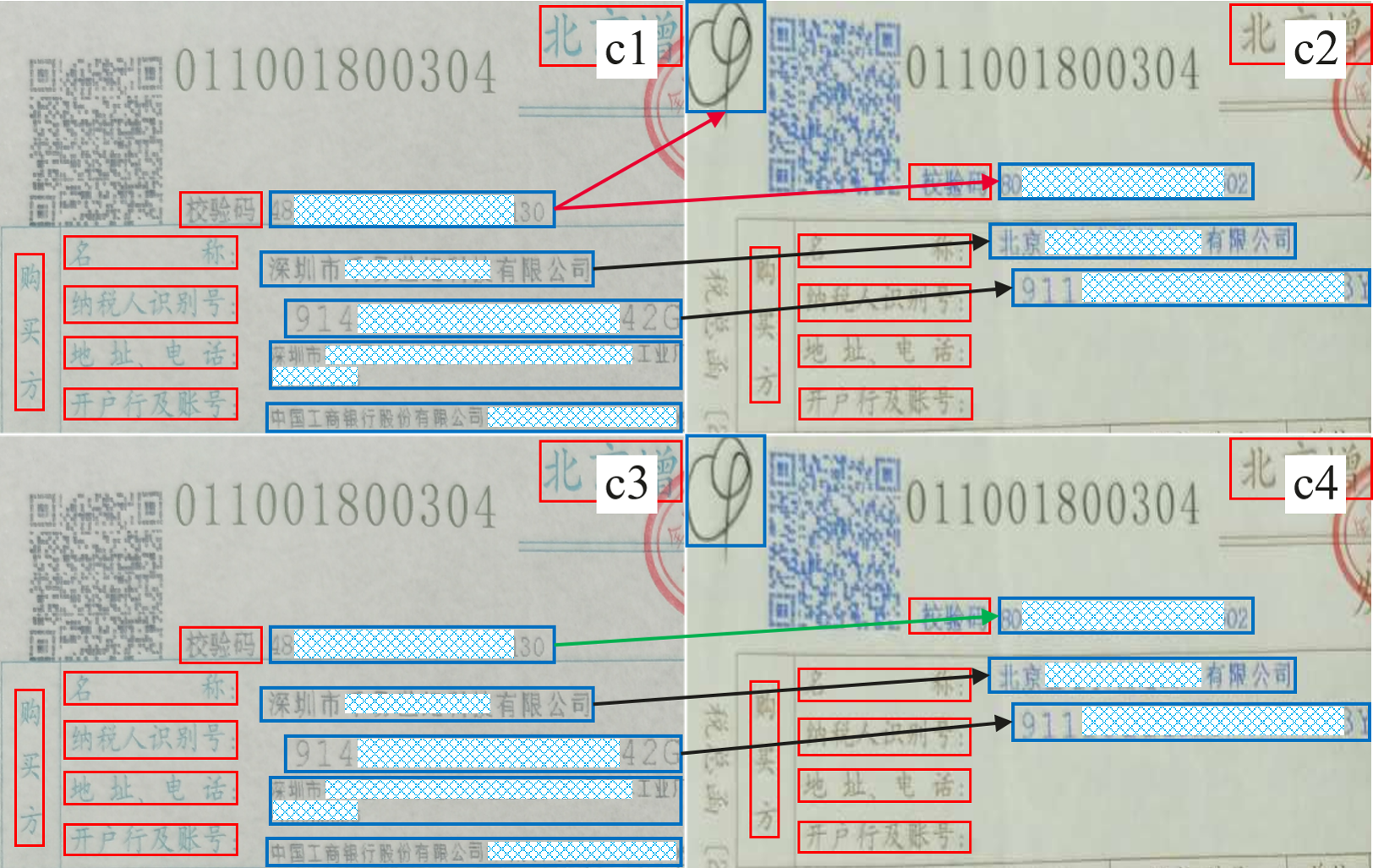}%
		\label{fig_first_case}}
% 	\hfil
	\caption{\textcolor{black}{
	The one-to-(at most)-one mapping constraint help resolve the problem of drifted fields and outliers. The red bounding boxes are landmarks, and the rest boxes are fields. We omit parts of boxes of both landmarks and fields for clearer illustration. The hand writen words in c2 and c4 are both ``$\textcircled{4}$".}}
	\label{fig:one_to_most_one}
\end{figure*}

\section{Our Model}\label{sec:methodology}
\textcolor{black}{In this section, we introduce our framework in detail. We present the framework of our model in Fig. \ref{fig:system}. In the first subsection, we define all the notations about graphs. In the second subsection, we discuss how to formulate the partial graph matching problem and how to annotate training data to avoid the many-to-many mapping, which violates the definition of graph matching problem~\cite{cheng2020one}, between fields. In the third subsection, we report important details of constructing graphs that consists of fields. In the forth subsection, we propose to use different MLP modules to calculate similarity scores between fields or edges based on different features. In the last subsection, we apply two solvers to the partial graph matching problem based on the similarity score.}

% 	We discuss cases 1, 2 and 3 in subfigure (a), and case 4 in subfigure (b). The black solid lines connecting support and query fields indicate the mapping between fields. We draw thicker lines in subfigure (a) because they correspond to cases 1, 2, and 3. In subfigure (b), we draw thinner lines because they are not related to case 4.

\subsection{Notations on Graphs}
We follow~\cite{hammami2015one} to call the set of dynamic text regions as \textbf{Fields}. \textcolor{black}{We use $f$ to note each field. We use the superscripts to differentiate the fields within one document, e.g., the $i$ th field is denoted as $f^i$.} Each field has its vertex features $x$ and label $y$. There are several ways to generate such node features. Firstly, we can use the set of static text regions, denoted as \textbf{Landmark} $L$, to generate the spatial feature of each field. Secondly, we can use text embedding to generate the semantic feature of each field. Thirdly, the aspect of each field, namely the width and height of its OCR bounding box, can also be a useful feature. For a document, we note the set of its fields as $F=\left\{f^i, 1<i<|F|\right\}$, the set of node features as $X=\left\{x^i\right\}$, and the set of  labels as $Y=\left\{y^i\right\}$. \textcolor{black}{Notations on edges are flexible. Given two fields $f^i,f^j\in F$, both $ij$ and $f^if^j$ can represent the directed edges from $f^i$ to $f^j$. We define the set of all edges to be $FF=\left\{ij,\forall f^i,f^j \in F\right\}$}. We can represent a document as a quaternion $G=\left\{F, FF, X, Y\right\}$.

We will use the subscripts to differentiate the support and query documents, i.e., $s$ represents the support document and $q$ represents the query document. \textcolor{black}{We use $f^i_q$ to represent the $i$ th field in a query document.} The one-shot KIE problem is to predict the label of each query field $f_{q}^{i}$ whose ground-truth label is $y_{q}^{i}$. We propose to solve the one-shot KIE problem using partial graph matching such that if the query field $f_{q}^i$ is matched with the support field $f_{s}^{a}$, then the model predict the label of $f_{q}^i$ to be $y_{s}^a$.
%However, there are several details to be handled carefully for us to solve the one-shot KIE problem using graph matching techniques. 

\subsection{Solving one-shot KIE with Partial Graph Matching}
Based on the above notations, the formulation of partial graph matching requires two additional concepts. \textcolor{black}{We follow Burkard et al.\cite{burkard1998quadratic} to use the concave quadratic formulation of the graph matching problem. Partial graph matching shares the same concepts but has different constraints.}  

The first one is the permutation matrix $P$, whose element $P_{ia}$ is $1$ if a query field $f_{q}^{i}$ is matched with a support field $f_{s}^{a}$, $0$ otherwise. This matrix describes the matching between $G_q$ with $G_s$, and has $|F_{q}|\times|F_{s}|$ elements. 

The second one is the affinity matrix $A$, which is a square matrix and operates on the vector version of $P$. Note that the vector version of $P$ is in the $\mathbb{R}^{|F_{q}|*|F_{s}|}$ space, and the shape of $A$ is $(|F_{q}|*|F_{s}|)\times(|F_{q}|*|F_{s}|)$. The elements of $A$ in different positions have different meanings. For the off diagonal elements, they describe how similar two edges are, where one edge comes from the graph $G_{q}$ and another one comes from $G_{s}$. If $f_{q}^{i}f_{q}^{j}$ is the edge in $FF_{q}$, and $f_{s}^{a}f_{s}^{b}$ is the edge in $FF_{s}$, then their similarity score in the affinity matrix $A$ is denoted as $A_{ij}^{ab}$. For the diagonal elements, we use $A_{ii}^{aa}$ to note how similar two fields are, i.e., $A_{ii}^{aa}$ is the similarity score between $f_{q}^{i}$ and $f_{s}^{a}$. 

Finally, the partial graph matching problem is formulated to be a constrained optimization problem, whose objective is:
\begin{alignat*}{2}
    \max_{P}\quad & \sum_{i=1}^{|F_q|}\sum_{a=1}^{|F_s|}P_{ia}A_{ii}^{aa}P_{ia} + \sum_{ij\in FF_q}^{|FF_{q}|}\sum_{ab\in FF_s}^{| FF_s|}P_{ia}A_{ij}^{ab}P_{jb}, & \tag{1}\\
    \mbox{s.t.}\quad& P\in\left\{P\mathbf{1}\leq\mathbf{1},P^{\top}\mathbf{1}\leq\mathbf{1},P\in\{0,1\}^{|F_{q}|\times|F_{s}|}\right\}, &\tag{2}
    \label{eqa:gm}
\end{alignat*}
where $\mathbf{1}$ is a column-wise vector whose elements are all one. \textcolor{black}{All the notations in equation (1) and (2) are fully explained in subsection B and A. The first inequality in equation (2) forbids a feasible permutation matrix $P$ to match multiple support fields with a target query field. The second inequality forbids $P$ to match multiple query fields with one support field. Both inequalities allow part of support and query fields to match with no fields. The first term in equation (1) sums over all possible matching between support and query fields to calculate the vertex similarity score. The second term sums over all possible matching between support and query edges to calculate the edge similarity score. A query edge $f_q^if_q^j$ is matched with a support edge $f_s^af_s^b$ if and only if both $f_q^i$ is matched with $f_s^a$ and  $f_q^j$ is matched with $f_s^b$.}

% \begin{figure}[t]
% 	\centering
% 	\includegraphics[width = 1.0\linewidth]{img/landmark_field_matching_illustration.pdf}\\
% 	\bcaption{\label{fig:one_to_most_one}The scenarios of fields matching. Better viewed in color. The boxes filled with color are landmarks, . The boxes has }
% \end{figure}

\textcolor{black}{Fig. \ref{fig:one_to_most_one} shows why the one-to-(at most) one constraint can be ensured, and how it helps resolve the problems of drifted fields and outliers. In practice, a document may contain many multi-line fields. The examples are the fields in subfigure (a). They are supposed to have the same label but are bounded by separate boxes. Cheng et al.~\cite{cheng2020one} suggests using the average boxes of support fields that share the same label to match with query fields. As shown in a1 and a2, their method leads to a one-to-many mapping that violates the one-to-(at most)-one mapping constraint. However, we propose to add number suffix to the label of multi-line support fields, so that the one-to-(at most)-one mapping between multi-region fields are possible as shown in a3 and a4. We remove the number suffix after prediction to restore the original label of each field.}

\textcolor{black}{In the subfigure (b), the yellow line segments indicate that the fields in b2 and b4 drifted towards down when compared with b1 and b3. We can see that a direct transport of spatial relationship from b1 to b2 failed. Particularly, the support field of ``$\$17.00$" is mapped to ``$\$0.00$" and ``$\$51.00$" at the same time. However, the one-to-(at most)-one constraint forbids our model to do so. In b2 and b4, to satisfy such constraint, our model choose to map each support field to the correct query field eventhough they are not the most similar field to each other in the aspect of spatial relationship.}
 
 \textcolor{black}{In the subfigure (c), c2 and c4 both contain the same outlier, which is ``$\textcircled{4}$". If there is no constraint, the method of Cheng et al~\cite{cheng2020one} will map a wrong support field to ``$\textcircled{4}$" as shown in c1 and c2. However, our model can refuse to match any support fields with ``$\textcircled{4}$" because of the constraint.}

\subsection{Document Graph Construction}
\noindent\textbf{Graph Vertices}. We only regard fields as graph vertices for both support and query documents. 

We use landmarks to generate spatial features for the fields. Specifically, for a target field, all the line segments connecting its center point with all landmarks will be arranged as a 2d matrix whose shape is $|L|\times2$. $|L|$ means the number of landmarks. The spatial features of different fields will be stacked such that the overall shape of one spatial features $X$ in a document is $|F|\times|L|\times2$. $|F|$ means the number of fields. We also use the OCR bounding box of each field to generate its aspect feature, i.e., the height and width of the bounding box are concatenated into a 2-dimensional feature. The aspect features in a document is of size $|F|\times2$. We use average word embedding to generate the textual features for each field. We use the pre-trained word embedding~\cite{P18-2023} with 300-dimension, and freeze it during training. The shape of the textual features in a document is with the size of $|F|\times300$.

For all documents, the landmarks and fields are detected by OCR systems automatically and then labeled manually. For each type of document, we will select one document as the support document, and the rest will serve as query documents. The support document should be as complete as possible. 

We will remove the extra landmarks for the query document and repair the missing ones compared to the support document. If a field is split into several parts because of an imperfect OCR system, then we will merge these fields. Note that this operation is possible only for the training data. The model will assign the ``outliers'' label to the extra fields during the evaluation process. 

\noindent\textbf{Graph Edges}. For each document, we build a visible graph among fields and then apply the Prime algorithm~\cite{cheriton1976finding} to get the minimum spanning tree of this graph. This tree is used as the final graph. Specifically, each field will emit 36 rays to search its visible neighbors. The resulting visible graph may contain many loops. We find that it is important to remove all the loops in the graph using the Prime algorithm in practice. The shorter edges connecting neighbor fields should be preserved to generate better performance. Each edge has two types of features: 1) The direction feature is the line segment connecting two fields. 2) We concatenate the height and width of the start field to the ones of the end field and generates a 4-dimensional feature as the aspect feature.

\begin{figure*}[t]
	\centering
	\subfloat[Training data, whose supervision signal is a permutation matrix.]{\includegraphics[width=0.49\linewidth]{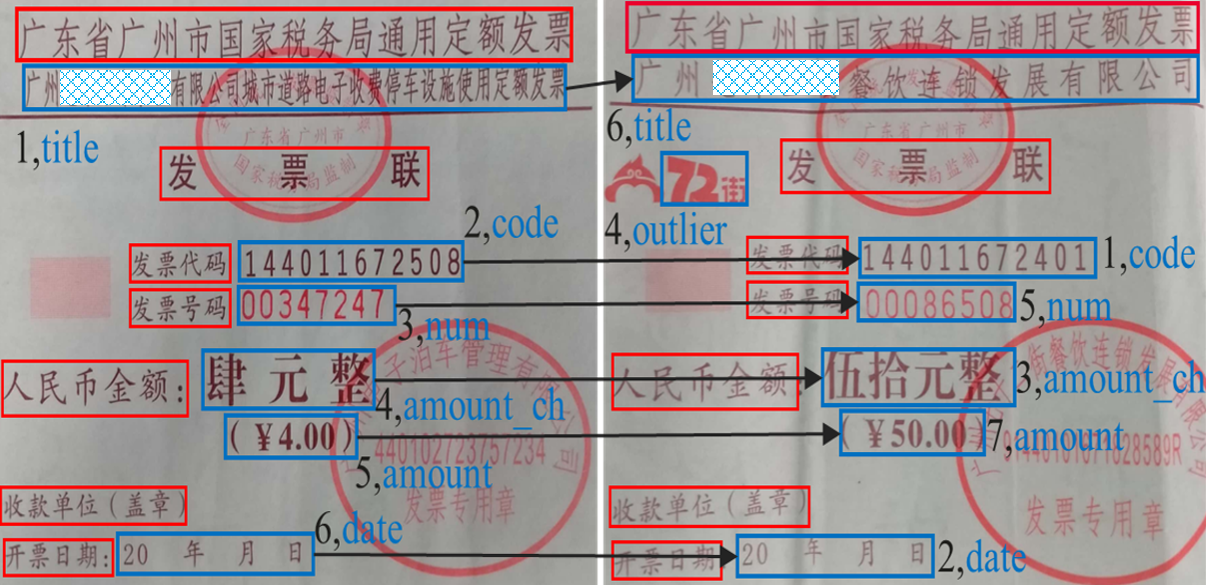}%
		\label{samples}}
	\hfil
	\subfloat[The permutation matrix.]{\includegraphics[width=0.238\linewidth]{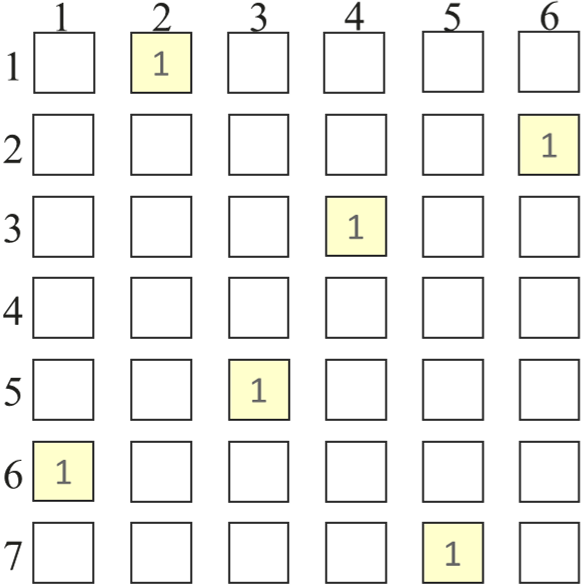}%
		\label{permut_matrix}}
	\hfil
	\subfloat[The vertex affinity matrix.]{\includegraphics[width=0.238\linewidth]{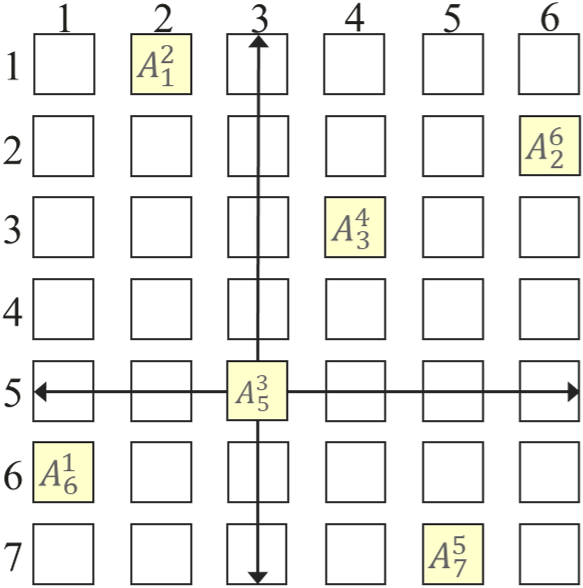}%
		\label{fig_second_case}}
	\caption{\textcolor{black}{In subfigure (a), the red boxes are landmarks, the blue boxes are fields, and the black lines indicate mapping from support to query fields. In subfigure (b), each row of the matrix corresponds to a query field, and each row has at most one entry being one with the other entries being zero.  The third entry of the fifth row is ``1", and this means that the fifth query field, whose text is ``00086508", corresponds to the third support field ``00347247". The fourth row has no entries being ``1", and this means that the forth query field has no matching support fields. We shuffle the index of query fields such that the permutation matrix will not be an identity matrix. In subfigure (c), similar to the permutation matrix, each row of the vertex affinity matrix corresponds to a query field. The entries are the similarity score, calculated by formula (\ref{eqa:vertex_affinity}), between support and query fields. We only show the entries of correct pairs of fields, and the other entries are not zero. The similarity score of correct pairs of fields should be larger than the wrong pairs that lie in the same row or column.}}
	\label{fig:ranking_loss}
\end{figure*}

\subsection{Vertex and Edge Affinities}
For a pair of fields, we can use multiple features to compute the vertex affinity between them. Specifically, we can compute their spatial, aspect, and textual affinities. Then, the average of these affinities is the final vertex affinity between them. We concatenate the features from query and support fields and then apply a Multi-layer Perceptron (MLP) to generate the affinity score between them. We will describe the process of computing spatial affinity matrix precisely. Aspect, textual, and edge affinity matrices are processed similarly.

For the field $f_q^i$ and $f_s^a$, their spatial features are $x_q^i$ and $x_s^a$. Both features are matrices with the same shape, $|L|\times 2$. We have aligned the landmarks such that the query documents always have the corresponding landmarks in the support document. Thus, it is reasonable to fix a specific landmark $l^k$ and then concatenate the two line segments $l^kf_q^i$ and $l^kf_s^a$. Note that $l^kf_q^i$ connects the center point of $f_q^i$ and $l^k$. $l^kf_s^a$ is similar. Let $l^kf_q^i\oplus l^kf_s^a$ be the concatenated feature, then the affinity score between $f_q^i$ and $f_s^a$ w.r.s to landmark $l^k$ is 
\begin{equation}
    Affi_{spatial}(f_q^i, f_s^a, l^k)=MLP(l^kf_q^i\oplus l^kf_s^a). \tag{3}
\end{equation}

By iterating over all landmarks, we can concatenate $x_q$ and $x_s$ into a $|L|\times 4$ matrix, which will be denoted as $x_q\oplus x_s$. Finally, the spatial affinity between $f_q^i$ and $f_s^a$ equals to the average of affinity score for all landmarks:
\begin{equation}
    A_{ii_{_{spatial}}}^{aa}=\frac{1}{|L|}\sum_{l^k=1}^{|L|}Affi_{spatial}(f_q^i, f_s^a, l^k). \tag{4}
\end{equation}
Note that the spatial affinity matrix $A_{ii_{_{spatial}}}^{aa}$ is calculated in a vectorized way, i.e., $X_q$ and $X_s$ are concatenated into a $|F_q|\times|F_s|\times|L|\times4$ tensor, then feed it into an MLP module to compute the affinity score tensor with the shape of $|F_q|\times|F_s|\times|L|$. We then average over the last dimension to obtain spatial affinity matrix $A_{ii_{_{spatial}}}^{aa}$ with the shape of $|F_q|\times|F_s|$. 

We compute the aspect, textual affinity matrices in a similar way but with separate MLP modules. The average of all affinity matrices is the final vertex affinity matrix computed as follow:
\begin{equation}
    A_{ii}^{aa}=\frac{1}{3}(A_{ii_{_{spatial}}}^{aa}+A_{ii_{_{aspect}}}^{aa}+A_{ii_{_{textual}}}^{aa}). \tag{5}
    \label{eqa:vertex_affinity}
\end{equation}

The off-diagonal elements of $A$ are calculated similarly. For edge $f_q^if_q^j$, we use $f_q^if_{q_{_{direct}}}^{j}$ to represent the direction feature, and use $f_q^if_{q_{_{aspect}}}^{j}$ to represent the aspect feature. Notations about the edge $f_s^af_s^b$ are similar. Then the similarity score between the two edges is computed by:
\begin{align}
A_{ij}^{ab}&=\frac{1}{2}(MLP(f_q^if_{q_{_{direct}}}^{j}\oplus f_s^af_{s_{_{direct}}}^{b}) \tag{6}\\
    &\quad +MLP(f_q^if_{q_{_{aspect}}}^{j}\oplus f_s^af_{s_{_{aspect}}}^{b})). \tag{7}
\end{align}
% \begin{equation}
%     A_{ij}^{ab}=\frac{1}{2}(MLP(f_q^if_{q_{_{direct}}}^{j}\oplus f_s^af_{s_{_{direct}}}^{b})+\\
%     &MLP(f_q^if_{q_{_{aspect}}}^{j}\oplus f_s^af_{s_{_{aspect}}}^{b})). \tag{6}
% \end{equation}

Note that we use separate MLP modules for vertex and edge affinities. 

\subsection{Combinatorial Solver}
Fig.\ref{fig:system} shows the pipeline of our model. We need to solve the partial graph matching problem and back-propagate through the solvers after calculating the affinity matrix.
                        
\noindent\textbf{Solving Partial Graph Matching Problem}.
Inspired by recent work of fusing deep learning and combinatorics\cite{vlastelica2019differentiation}, we adopt two solvers to solve the partial graph matching problem. The first solver is DD-ILP solver~\cite{swoboda2017dual}, which is a third party libraries that aim to solve a specific type of discrete optimization problem called Integer-Relaxed Pairwise-Separable Linear Programs (IRPS-LP), and the partial graph matching problem with formulation (1) and (2) is an example of such problems. 

\textcolor{black}{We also reimplement the ZAC-GM solver in~\cite{wang2020zero}. Although the formulation of the ZAC-GM solver is not the same as equations (1) and (2), the input, output, and constraints of this solver are the same as DD-ILP. In~\cite{wang2020zero}, the authors clarify a sufficient condition about when a vertex affinity matrix can lead to an optimal permutation matrix that represents the correct mapping between vertexes. Inspired by this sufficient condition, we design an additional ranking loss to regularize the MLP modules. We use this ranking loss to enlarge the similarity score difference between correct vertex pairs and the wrong vertex pairs during training.}

\textcolor{black}{Fig. \ref{fig:ranking_loss} illustrates how to calculate this ranking loss for a pair of support and query documents. Take the support field ``00347247" and the query field ``00086508" for example. Their similarity score should be higher than the score between ``00347247" and any other query fields. Their score should be higher than the score between ``00086508" and any other support fields too. Experiments show the effectiveness of this ranking loss.}

% The solvers will generate a permutation matrix $\hat{P}$, which satisfies the one-to-(at most)-one constraint as described in equation (\ref{eqa:gm}). There are several advantages to adopt such solvers. Firstly, their implementation is highly optimized. Secondly, the one-to-(at most)-one mapping constraint can ensure that the drifted fields will be assigned to the correct labels. Fields drift are common in printed document, as shown in Fig.~\ref{fig:spatial_drift}(b), the field \textit{balance} is printed far away from its corresponding landmark compared with the support document in ~\ref{fig:spatial_drift}(a). The mapping constraint can help ease this problem. In general, the solvers will generate globally optimized solutions.

\noindent\textbf{Back Propagate Through Solver}. We adopt the hamming loss between the predicted $\hat{P}$ and the label $P^*$. A fundamental problem of fusing deep learning and combinatorics is that the gradient of neural networks tends to be zero most times. In our model, a subtle change of the affinity matrix will not change the predicted permutation matrix $\hat{P}$, i.e., $\hat{P}$ is a piece-wise constant function w.r.t the parameters of MLP modules. To overcome this problem, we adopt the techniques described in \cite{rolinek2020deep}. The additional benefits of using hamming loss are that the wrong prediction can also generate gradients as shown in Fig. \ref{fig:system}, this leads to a faster convergence compared with the cross-entropy (CE) loss in the LF-BP model. For example, the CE loss will only consider the negative diagonal elements in $P^*$ in Fig.~ \ref{fig:system}, while the hamming loss will also propagate through those non-zero and off-diagonal elements.

% \begin{table}[!hbt]
% \caption{Statistics of DKIE train datasets.}
% \label{tab:dataset_statistics}
% \centering
% \begin{tabular}{c c c c c}
%     \hline
%     \hline
%      Dataset & Description & $\#$ Styles & $\#$ Docs &  $\#$ Fields \\
%      \hline
%      d0 & Taxi receipts & 7 & 300 & 13 \\
%      \hline
%      d3 & PT invoice (Special) & 2 & 138 & 15 \\
%      \hline
%      d4 & VAT invoice (Normal) & 2 & 138 & 12 \\
%      \hline
%      d5 & Ferry tickets & 1 & 100 & 14 \\
%      \hline
%      d6 & Airline itinerary & 2 & 138 & 25 \\
%      \hline
%      d7 & VAT invoice (Special) & 2 & 157 & 44 \\
%      \hline
%      d8 & Medical invoice & 2 & 164 & 36 \\
%      \hline
%      \hline
% \end{tabular}
% \end{table}

\begin{table}[!hbt]
\caption{Statistics of DKIE datasets.}
\label{tab:dataset_statistics}
\centering
\begin{tabular}{c c c c c}
    \hline
    \hline
     Dataset & Description & $\#$ Styles & $\#$ Docs &  $\#$ Fields \\
    %  \hline
    %  d0 & Taxi receipts & 12 (7/5) & 356/1741 & 16 \\
    \hline
     \textcolor{black}{d0} & Taxi receipts & 12 (7:5) & 136 & 16 \\
     \hline
     \textcolor{black}{d1} & CHSR tickets & 1 (All test) & 169 & 11 \\
     \hline
     d2 & Bording pass & 2 (1:1) & 54 & 10 \\
     \hline
     d3 & PT invoice (Special) & 2 (1:1) & 151 & 15 \\
     \hline
     \textcolor{black}{d4} & VAT invoice (Normal) & 2 (1:1) & 118 & 12 \\
     \hline
     d5 & Ferry tickets & 2 (1:1) & 98 & 14 \\
     \hline
     d6 & Airline itinerary & 3 (2:1) & 107 & 25 \\
     \hline
     \textcolor{black}{d7} & VAT invoice (Special) & 2 (1:1) & 118 & 44 \\
     \hline
     d8 & Medical invoice & 3 (2:1) & 163 & 36 \\
     \hline
     \textcolor{black}{d9} & Quota invoice & 4 (All train) & 162 & 9 \\
     \hline
     d10 & Bank card & 1 (All train) & 197 & 8 \\
     \hline
     d11 & Express bill & 1 (All train) & 157 & 5 \\
     \hline
     d12 & Toll fee & 1 (All train) & 151 & 10 \\
     \hline
     d13 & Customs declaration & 3 (All train) & 158 & 14 \\
     \hline
     d14 & Duty-paid proof & 3 (All train) & 106 & 6 \\
     \hline
     d15 & Car-hailing receipts & 2 (All train) & 162 & 21 \\
     \hline
     d16 & VAT invoice (Volume) & 2 (All train) & 151 & 33\\
     \hline
     \hline
\end{tabular}
\end{table}

% \newpage
\begin{table*}[!hbt]
    \centering
    \bcaption{Comparison with state-of-the-art supervised and one-shot learning methods.}
    \label{tab:over_all_results}
    \begin{tabular}{c|c|c|c|c|c|c|c|c|c|c|c|c|c}
    \hline
    \hline
     \multicolumn{2}{c|}{\multirow{2}{*}{Method}} & \multirow{2}{*}{Features} & Training & Testing & \multicolumn{8}{c}{Accuracy(\%)}\\ 
    \cline{6-14}
    \multicolumn{2}{c|}{} & & data size & data type & d0 & d1 & d2 & d3 & d4 & d5 & d6 & d7 & d8\\ 
    \hline
    % Supervised & LayoutLM~\cite{xu2020layoutlm} & spatial+visual & \multirow{2}{*}{3K per style} & \multirow{5}{*}{clean}  & - & - & - & - & - & - & - & - & - \\
    % \cline{2-2}
    % \cline{6-14}
    % learning & PICK~\cite{Yu2020PICKPK} & +text & & & - & - & - & - & - & - & - & - & - \\
    % \cline{1-4}
    % \cline{6-14}
    \multirow{3}{*}{One-shot} & LF-BP~\cite{cheng2020one} & \multirow{3}{*}{spatial} & \multirow{3}{*}{1861 all styles} & \multirow{3}{*}{clean} & 94.7 & \textcolor{black}{99} & 100 & 100 & 92.1 & \textcolor{black}{100} & \textcolor{black}{92.6} & \textcolor{black}{100} & \textcolor{black}{100}\\
    \cline{2-2}
    \cline{6-14}
    & Ours (DD-ILP) & & & & \textcolor{black}{99.1} & \textcolor{black}{100} & \textcolor{black}{100} & \textcolor{black}{100} & \textcolor{black}{100} & \textcolor{black}{100} & \textcolor{black}{98} & \textcolor{black}{100} & \textcolor{black}{100}\\
    \cline{2-2}
    \cline{6-14}
    & Ours (ZAC-GM) & & & & 99.1 & \textcolor{black}{100} & 100 & 100 & 100 & \textcolor{black}{100} & \textcolor{black}{97.8} & \textcolor{black}{100} & \textcolor{black}{100}\\
    \hline
    % Supervised & LayoutLM~\cite{xu2020layoutlm} & spatial+visual & \multirow{2}{*}{3K per style} & \multirow{5}{*}{drifted}  & - & \multirow{5}{*}{-} & - & - & \multirow{5}{*}{-} & - & - & \multirow{5}{*}{-} & - \\
    % \cline{2-2}
    % \cline{6-6}
    % \cline{8-9}
    % \cline{11-12}
    % \cline{14-14}
    % learning & PICK~\cite{Yu2020PICKPK} & +text & & & - &  & - & - &  & - & - &  & - \\
    % \cline{1-4}
    % \cline{6-6}
    % \cline{8-9}
    % \cline{11-12}
    % \cline{14-14}
    \multirow{3}{*}{One-shot} & LF-BP~\cite{cheng2020one} & \multirow{3}{*}{spatial} & \multirow{3}{*}{1861 all styles} & \multirow{3}{*}{drifted} & 60 &  & 68.6 & 42.4 &  & \textcolor{black}{80.9} & \textcolor{black}{75.6} &  & \textcolor{black}{65.2}\\
    \cline{2-2}
    \cline{6-6}
    \cline{8-9}
    \cline{11-12}
    \cline{14-14}
    & Ours (DD-ILP) & & & & \textcolor{black}{97} & - & \textcolor{black}{71.1} & \textcolor{black}{90} & - & \textcolor{black}{100} & \textcolor{black}{96} & - & \textcolor{black}{96.2}\\
    \cline{2-2}
    \cline{6-6}
    \cline{8-9}
    \cline{11-12}
    \cline{14-14}
    & Ours (ZAC-GM) & & & & 96.3 &  & 71.1 & 93.9 &  & \textcolor{black}{100} & \textcolor{black}{96} &  & \textcolor{black}{96.2}\\
    \hline
    % Supervised & LayoutLM~\cite{xu2020layoutlm} & spatial+visual & \multirow{2}{*}{3K per style} & \multirow{5}{*}{outliers}  & - & - & - & - & - & - & - & - & - \\
    % \cline{2-2}
    % \cline{6-14}
    % learning & PICK~\cite{Yu2020PICKPK} & +text & & & - & - & - & - & - & - & - & - & - \\
    % \cline{1-4}
    % \cline{6-14}
    \multirow{3}{*}{One-shot} & LF-BP~\cite{cheng2020one} & \multirow{3}{*}{spatial} & \multirow{3}{*}{1861 all styles} & \multirow{3}{*}{outliers} & 64.4 & \textcolor{black}{97.9} & 90 & 81.3 & 93.1 & \textcolor{black}{97} & \textcolor{black}{70} & \textcolor{black}{91.2} & \textcolor{black}{90}\\
    \cline{2-2}
    \cline{6-14}
    & Ours (DD-ILP) & & & & \textcolor{black}{89} & \textcolor{black}{97.2} & \textcolor{black}{85.2} & \textcolor{black}{79.1} & \textcolor{black}{86.3} & \textcolor{black}{96.2} & \textcolor{black}{88} & \textcolor{black}{91.3} & \textcolor{black}{95.8}\\
    \cline{2-2}
    \cline{6-14}
    & Ours (ZAC-GM) & & & & 91.2 & \textcolor{black}{98.7} & 90 & 100 & 98 & \textcolor{black}{100} & \textcolor{black}{95.8} & \textcolor{black}{96} & \textcolor{black}{97}\\
    \hline
    Supervised & LayoutLM~\cite{xu2020layoutlm} & spatial+visual & \multirow{2}{*}{3K per style} & \multirow{5}{*}{all}  & 97.3 & 97.0 & - & 94.6 & 95.0 & 96.1 & 91.8 & 94.3 & 92.5 \\
    \cline{2-2}
    \cline{6-14}
    learning & PICK~\cite{Yu2020PICKPK} & +text & & & 97.9 & 97.8 & - & 95.8 & 95.7 & 96.6 & 92.3 & 94.9 & 92.2 \\
    \cline{1-4}
    \cline{6-14}
    \multirow{3}{*}{One-shot} & LF-BP~\cite{cheng2020one} & \multirow{3}{*}{spatial} & \multirow{3}{*}{1861 all styles} & & 80.39  & 98.2 & 84.1 & 94.4& 92.5& 97.3& 87.2 & 95.8 &93.8\\
    \cline{2-2}
    \cline{6-14}
    & Ours (DD-ILP) & & & & 93.6 & 98.5 & 84.7 & 96.0 & 97.1 & 98.4 & 94.4 & 96.1 & 97.2\\
    \cline{2-2}
    \cline{6-14}
    & Ours (ZAC-GM)  & & & & 95.7 & \textcolor{black}{99} & 85.2 & 98.5 & 99.5 & \textcolor{black}{100} & \textcolor{black}{96.4} & 97.2 & \textcolor{black}{98}\\
    \hline
    \hline
    \end{tabular}
\end{table*}

\section{EXPERIMENTS}\label{sec:experiments}
\subsection{Datasets}
\textcolor{black}{The datasets of DocLL/SROIE-Oneshot collected by \cite{cheng2020one} are not released by the day we submit our paper. Therefore, we created our own datasets to promote the research in the one-shot KIE task, especially on the problems of drifted fields and outliers. To generate a fair comparison on our datasets in this paper, we used the same features and training settings as \cite{cheng2020one}. }

\noindent\textbf{DKIE One-shot Dataset}. We created a new dataset consisting of 2,500 documents. \textcolor{black}{We report the statistics of the DKIE dataset in Table \ref{tab:dataset_statistics}. The dataset can be grouped into 17 big categories such as Value-Added Tax (VAT) invoices, Medical invoices, and Taxi receipts. Each category may contain different styles of documents. Different styles of documents in one category will have similar layouts. However, each style needs independent support document because different styles have different landmarks and labels for the one-shot learning methods.}  Fig.\ref{fig:our_dataset} shows sample images from the testing set. The dataset is challenging because the document images are taken by smartphones. Thus, 3D distortions, variant image size, drifted fields, and unmatched fields are quite common.

\noindent\textbf{Groundtruth Generation}. We asked two annotators to label the data separately. We cross-checked and rectified the incorrect labels. We repair the missing landmarks with dummy bounding boxes for a query document during the inference process to guarantee the support and query have the same number of landmarks. For multi-region fields, we add number suffix to the original labels as suggested in the B subsection of our model.

\subsection{Implementation Details}	
\noindent\textbf{State-Of-The-Art models}
We compared our method with the LayoutLM\cite{xu2020layoutlm}, PICK\cite{Yu2020PICKPK}, and LF-BP\cite{cheng2020one}. Layout LM model and the PICK model are supervised-learning models. LM-BP model and our model are one-shot learning models.

\noindent\textbf{Training Details}. 
We used Pytorch to implement our models. All the models are trained on one NVIDIA Tesla V100 GPU with 16G memory. We applied ADAM to optimize the model with a batch size of 8 and trained the model on a single GPU card. \textcolor{black}{The initial learning rate is 0.05 and decays by 0.85 after each epoch.} 

% The training process takes around an hour for each epoch and the model converged after five epochs.

To keep good performance, supervised-learning-based models generally maintain different parameters for different styles of documents. \textcolor{black}{Therefore, we need to split each style of documents into training and testing data to train separate parameters for each style. We generate 3,000 documents for each style, including all the real samples, and the rest are synthesized.} The rest training details are the same as the original methods discussed in \cite{xu2020layoutlm,Yu2020PICKPK}.
% 如何保证监督学习的方法和one-shot的方法在同一批数据上测试。

%Therefore, we construct separate training data for different styles of documents.

On the contrary, one-shot-learning-based models can handle different styles of documents using the same parameters. \textcolor{black}{Therefore, we should train one model with different styles of documents. For one-shot learning methods, We split the documents of each category into training and testing sets according to their styles. For example, there are 12 styles in the taxi receipts category. We choose 7 styles as the training data and the rest 5 styles as the testing data. The third column of Table \ref{tab:dataset_statistics} shows the number of styles of each category. This column also shows the ratio between training and testing styles in the parentheses. The number of all training documents is 1861 and the number of all testing documents is 639.}

% One-shot learning based models save much labor since that users only need to annotate one sample for each style of documents. 
% We select 2000 documents for training, and those types of training data will not overlap with the test documents.

\noindent\textbf{Testing Details}. For one-shot-learning-based models, we predefined the support document of each style, and then different images of the same style will serve as the query documents. \textcolor{black}{For each style, a document containing as many landmarks/fields as possible serves to be a good support document.} Our model will predict the label of each field in query documents using those labels defined in the support documents. 

\textcolor{black}{To study the performance of our model on samples containing drifted fields and outliers separately, we further split the testing data of each style into 3 parts as shown in the fifth column of Table \ref{tab:over_all_results}. There are ``clean" documents in which all fields are nicely aligned with the landmarks and have no outliers. There are ``drifted" documents in which some fields are drifted so badly that even humans need to check each field very carefully to judge the labels. There are also documents containing ``outliers". A small number of documents contain both drifted fields and outliers. We include them in ``drifted" and ``outliers" datasets at the same time. In the ``all" dataset, we report the average accuracy of fields in all query documents within the same category.}

\subsection{Experimental Results}\label{sec:experimental_results}
In this section, we compared our method with existing SOTA methods. The results are presented in Table~\ref{tab:over_all_results}. For a fair comparison, we only use spatial features in our model because the LF-BP model~\cite{cheng2020one} only consumes spatial features. The effect of other features will be discussed in Section~\ref{sec:ablation_study}.

% The supervised-learning-based models do not have advantages over the one-shot-learning-based models on the clean data despite that the one-shot-learning-based models require fewer labeled data. However, o
\subsubsection{Performance on ``Clean" Documents}
\begin{figure}[h]
	\centering
	\includegraphics[width = 1.0\linewidth]{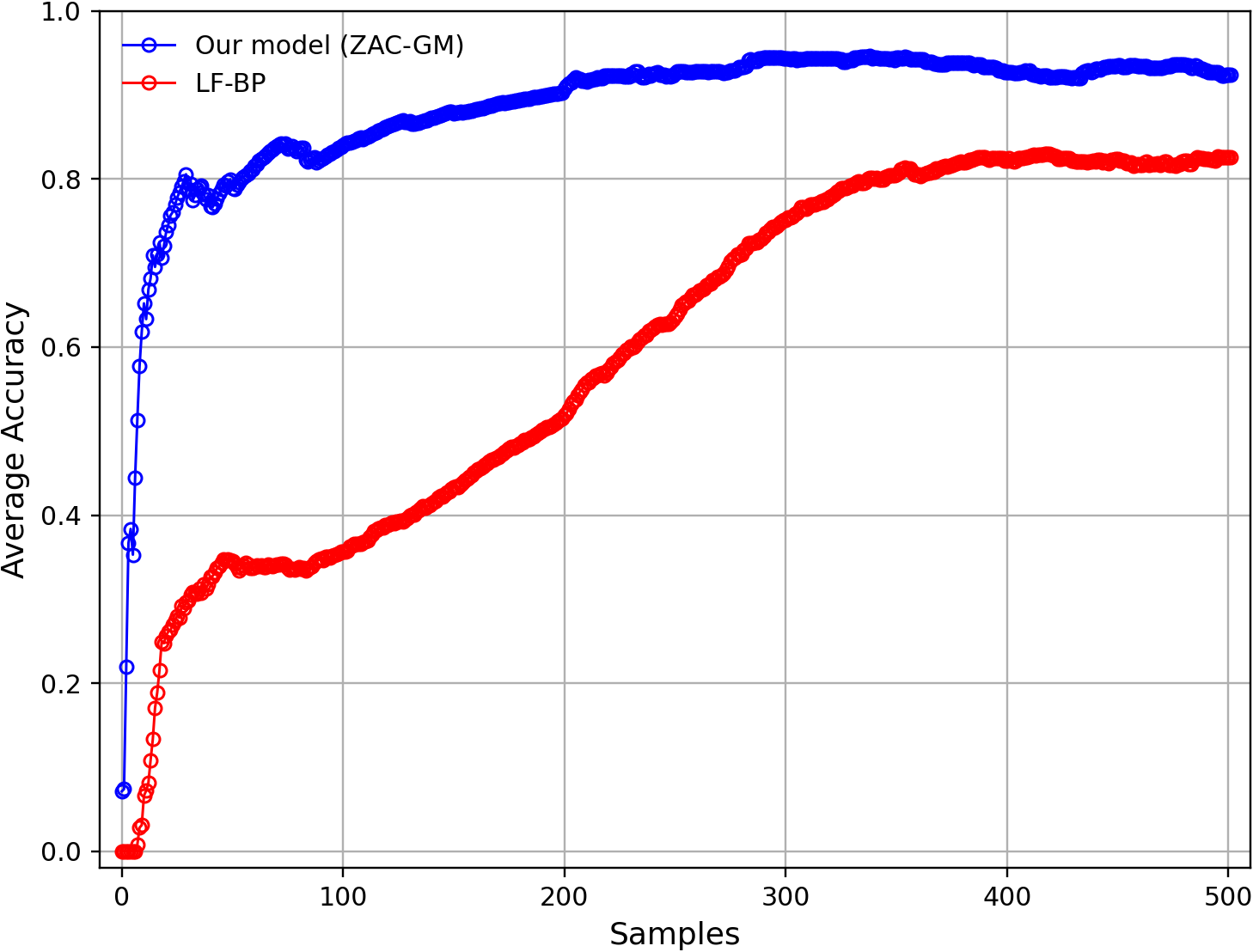}
	\caption{\label{fig:average_acc}Average accuracy of all fields from different documents.}
\end{figure}

\textcolor{black}{All models perform good on the ``clean" data in Table~\ref{tab:over_all_results}. Our model converges faster than the LF-BP model in the training stage. For each iteration, we calculate the average accuracy of fields in all training documents. We show part of the increasing process in Figure~\ref{fig:average_acc}. Different from the LF-BP model whose average accuracy increased relatively smoothly, the accuracy of our model increased drastically in the initial training stage. There are two reasons explaining the rapidly increasing process of accuracy of our model. First, our model uses hamming loss to calculate the gradients while the LF-BP model uses the cross entropy loss. For each query field, when our model matches it with the wrong support field, the hamming loss will generate gradients not only based on the labeled pairs of fields but also the wrong pairs. In the contrary, the cross entroy loss will only generate gradients based on the labeled pairs of fields. Second, the combinatorial solvers in our model are not sensitive to the suttle change of affinity matrices, which are the outputs of MLP modules. Therefore, MLP modules in our model only need to output roughly correct affinity matrices such that the combinatorial solvers can find the correct mapping between support and query fields. This also leads to a faster increasing process of accuracy.}

\subsubsection{Performance on ``Drifted" Documents}
\begin{figure}[t]
	\centering
	\includegraphics[width = 1.0\linewidth]{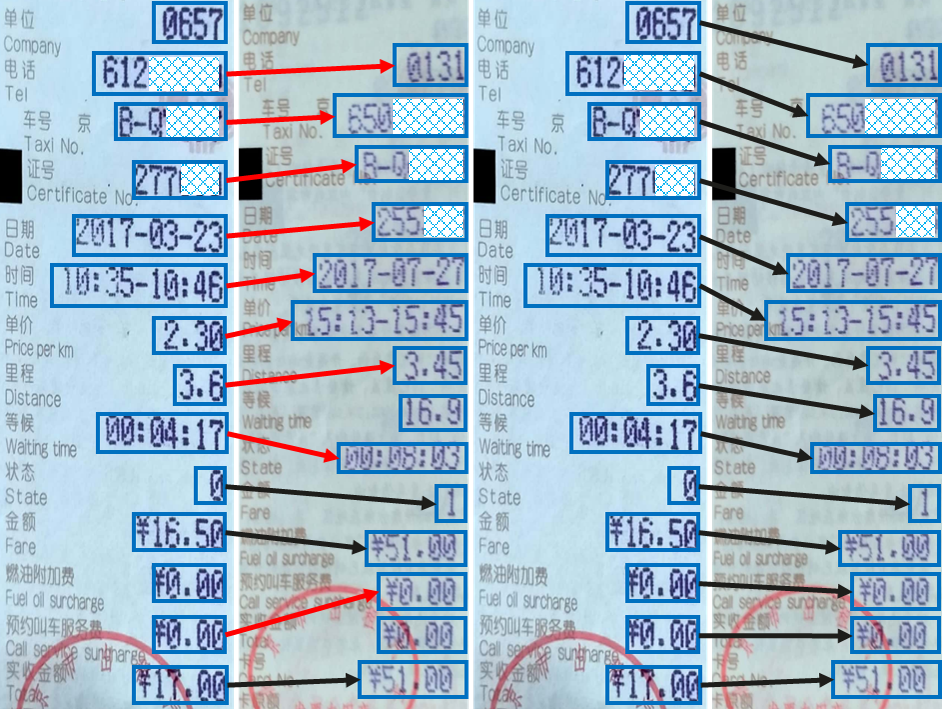}
	\caption{\label{fig:drifted_data} LF-BP (left) and our model's (right) prediction about documents containing drifted fields in the d0 dataset. LF-BP model failed while our model can handle the drifted fields.}
\end{figure}

\textcolor{black}{The LF-BP model failed on the ``drifted" data in Table~\ref{tab:over_all_results} while the performance of our model droped moderately. Our model significantly outperform the LF-BP model accross all datasets. Especially on the d0 dataset. We find that the fields in this dataset aranged vertically. If one of the fields in the head part of a document drifted downwards, then all the fields below it will also drifted downwards. Typical samples of d0 dataset can be found in Figure~\ref{fig:spatial_drift}, Figure~\ref{fig:one_to_most_one} (b) and Figure~\ref{fig:drifted_data}. Both the online demo released by~\cite{cheng2020one} and our reimplemented LF-BP model achieve low accuracy on the drifted fields. We show typical mistakes made by the LF-BP model in Figure~\ref{fig:drifted_data}. The left pair of documents in Figure~\ref{fig:drifted_data} shows the prediction of the online demo released by~\cite{cheng2020one}. Red lines indicate wrong mapping between fields. Although the LF-BP model does not map ``16.9" and ``\$0.00" to any support fields, the authors of \cite{cheng2020one} do not report this feature in their paper. We analysis why the LF-BP model failed on the drifted fields in the case study section~\ref{subsec:case_study} using the samples in the d0 dataset.}

\begin{figure*}[!htb]
	\centering
	\includegraphics[width = 1.0\linewidth]{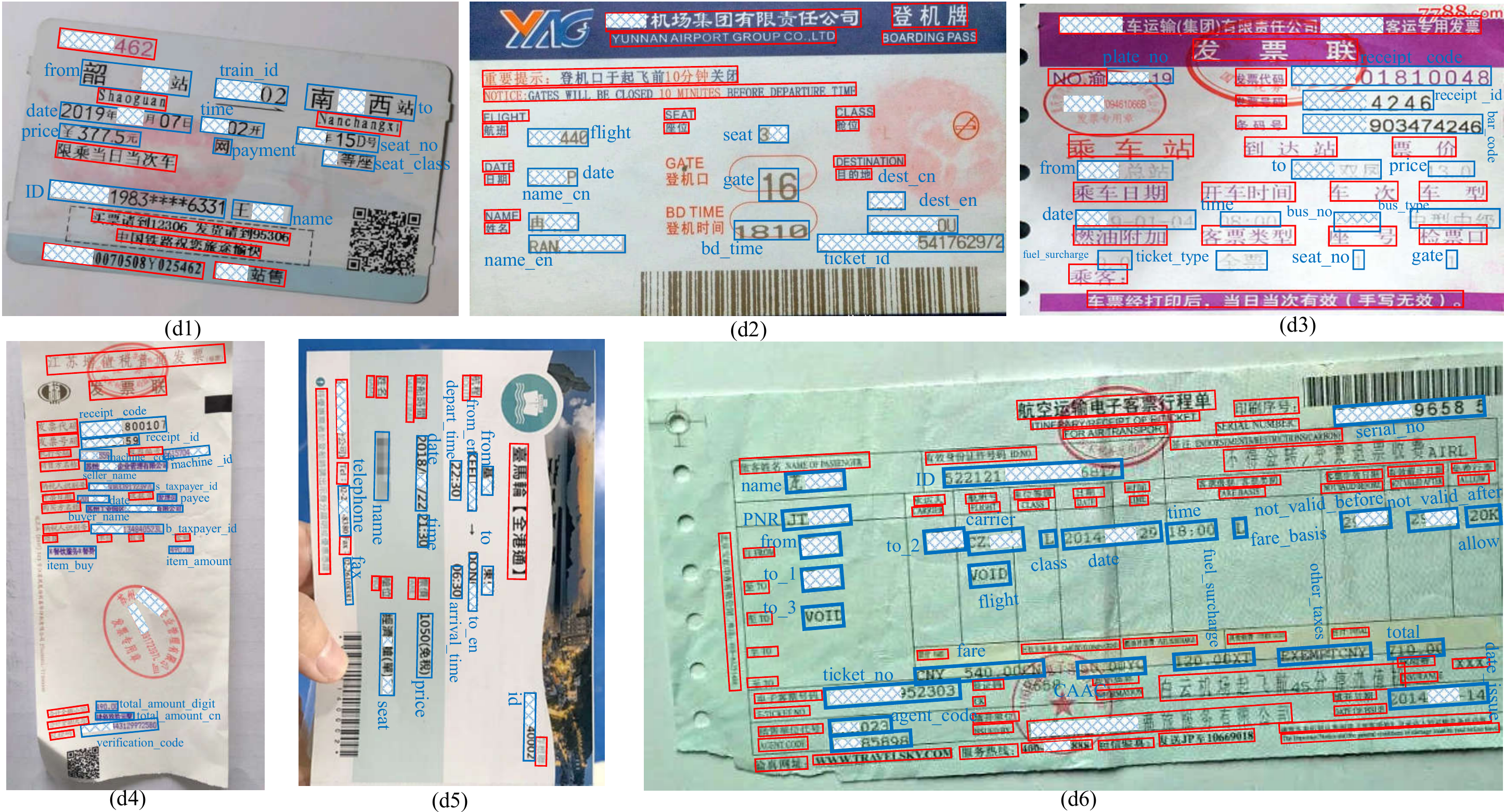}\\
	\includegraphics[width = 1.0\linewidth]{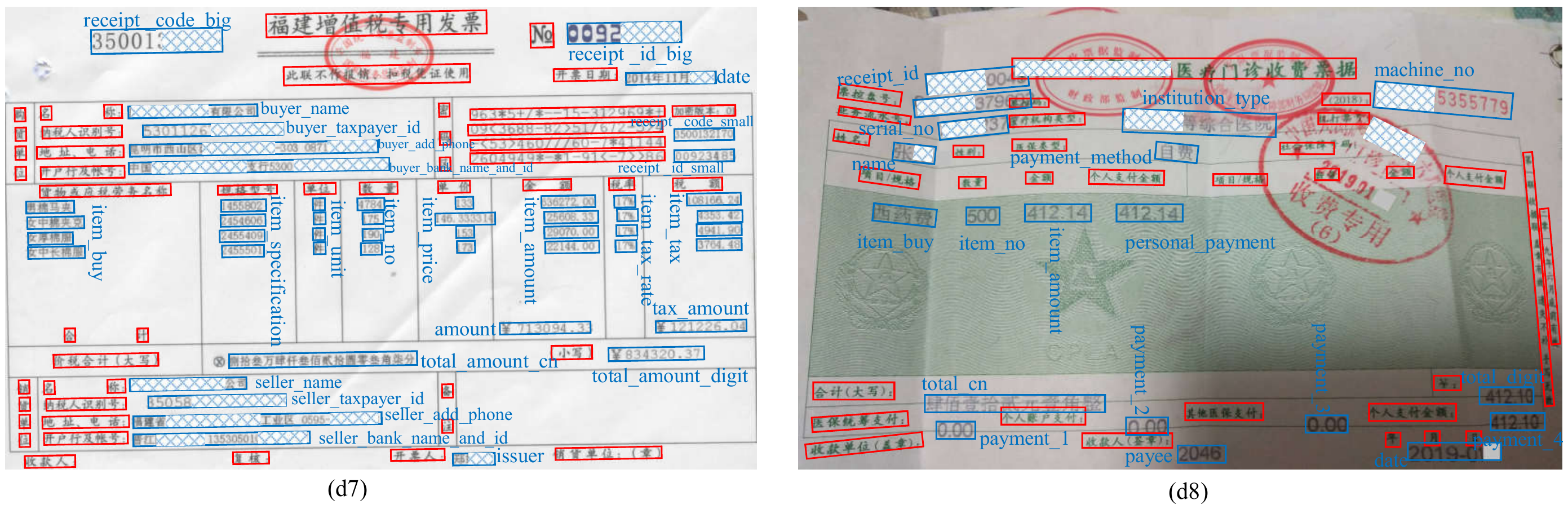}\\
	\bcaption{\label{fig:our_dataset}Samples in test dataset of DKIE. The sensitive information has been masked.}
\end{figure*}

\textcolor{black}{Notice that d1, d4 and d7 do not have documents that contain drifted fields. The fields drifted horizontally in d3, d5, d6 and d8 datasets. Around half of documents in the d2 dataset contain flipped fields as shown in Figure~\ref{fig:flipped_fields}. The models can not predict the labels of these flipped fields correctly solely based on the spatial features. However, our model with ZAC-GM solver can achieve good performance when it use additional features such as the width and height of bounding boxes of fields. We discuss this problem in the subsection~\ref{sec:ablation_study}. Figure~\ref{fig:our_dataset} shows examples of drifted fields in all datasets.}

\subsubsection{Performance on ``Outliers" Documents}
\textcolor{black}{Table~\ref{tab:over_all_results} shows that our model, when using ZAC-GM solver, is the only one succeed across all datasets. When our model use the DD-ILP solver, it can not handle documents containing outliers. We investigate the original code and find that it aims to solve the graph matching problem and requires the support and query documents to have the same number of fields. This is not true in documents containing outliers. However, when we reimplement the ZAC-GM solver~\cite{wang2020zero} and employ it to pick out the outliers, our model can handle the drifted fields and outliers at the same time to some extend. By taking a close look at the prediction of our reimplemented LF-BP model, we find that the outliers are predicted to have a label of their neighbors as shown in the Figure~\ref{fig:one_to_most_one} (c1) and (c2). Despite that~\cite{cheng2020one} did not report how they handle the outliers, the demo released by them will refuse to output the label of some outliers. We argure that our model provides an alternative approach that can handle the outliers.} 

\begin{figure}[t]
	\centering
	\includegraphics[width = 1.0\linewidth]{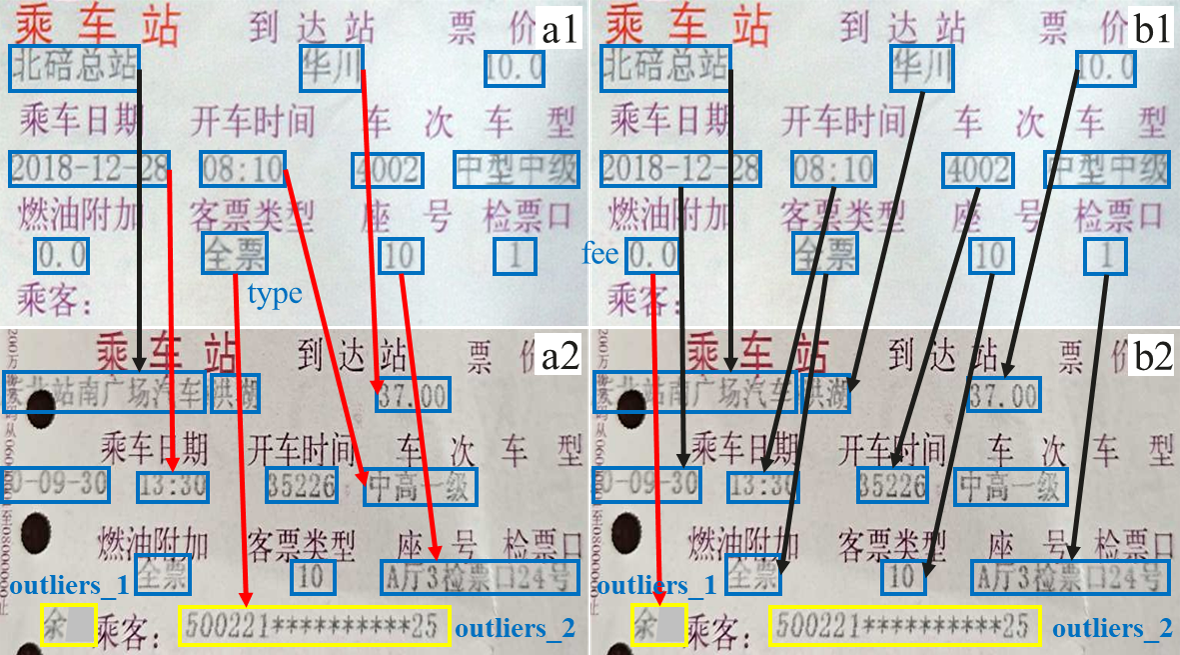}
	\caption{\label{fig:d3_compare} Documents containing both drifted fields and outliers in the d3 dataset. (a1) and (a2) show the prediction of LF-BP model. (b1) and (b2) show the prediction of our model with ZAC-GM solver. Yellow boxes are outliers. Red lines indicate wrong mapping between fields. Both models failed on this pair of documents.}
\end{figure}

\textcolor{black}{Some documents in the d0, d3 and d6 dataset contain drifted fields and outliers at the same time. We find that such documents are the most difficult ones. Not only the LF-BP model failed on these documents, but also the performance of our model dropped by a relatively large margin. Figure~\ref{fig:d3_compare} shows such samples in the d3 dataset. We find that when the outliers are close to the drifted fields, they are hard to distinct with each other solely based on their spatial features. For example, LF-BP model maps the ``type" field in (a1) to the ``outliers$\_$2" field in (a2). Our model also maps the ``fee" field in (b1) to the ``outliers$\_$1" field in (b2). The position of these outliers are so close to other drifted query fields such that the models may confuse them with the situation of multi-region fields. This indicates that we should measure the imilarity between fields using more diverse features such as the width and height of bounding boxes of fields or the text embedding in fields. We discuss the impact of different features in the ablation study section~\ref{sec:ablation_study}.}

\subsubsection{Performance on ``All" Documents}
Our model outperformed the LF-BP method on all datasets. This is because our datasets are more challenging. There are many documents containing spatial drifted fields and outliers. In contrast to the method of LF-BP, the solving process of our model generates globally optimized results through the one-to-(at most)-one constraint. Such constraint overcomes the difficulties brought by the spatial drifted fields and outliers. Our model achieved competitive performance against the supervised-learning-based models on the d0 dataset, and better results on the other testing datasets. This result is not surprising because the supervised-learning-based models have three unfair advantages. First, they consume a lot more labeled documents than one-shot-learning-based methods during the training stage, namely 3,000 samples per style versus 1861 samples for all styles. Second, they need to train separate models for different types of documents, which can help them fit the data bias in each type of document. In contrast, one-shot learning based models use one model for all types of documents. Therefore, for each new style of document, our model benefits from the other styles of documents and requires only one labeled sample to serve as the support document for each type. Lastly, they consume more features than the one-shot-learning-based models in this experiment, namely spatial, visual, and text features versus spatial feature only.

\begin{figure}[t]
	\centering
	\includegraphics[width = 1.0\linewidth]{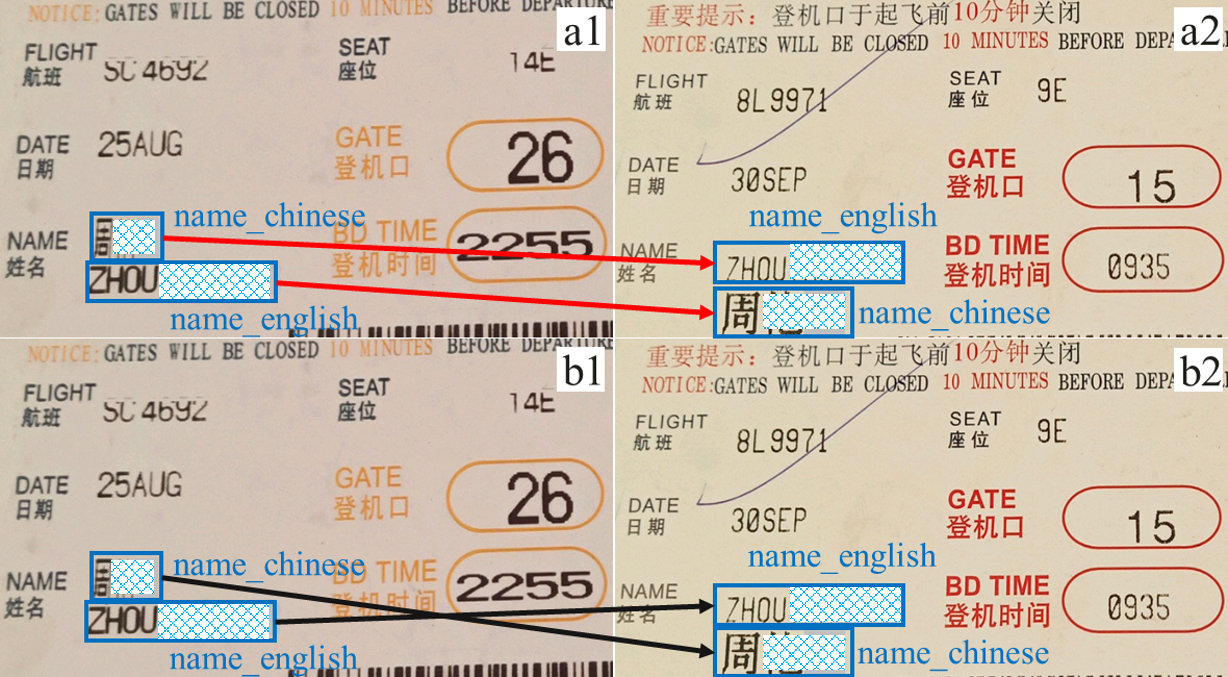}
	\caption{\label{fig:flipped_fields} Documents containing both flipped fieldsin the d2 dataset. (a1) and (a2) show the prediction of our model without using aspect features. (b1) and (b2) show the prediction of our model using aspect features. Red lines indicate wrong mapping between fields.}
\end{figure}

\begin{table}[t]
\bcaption{Aspect features help deal with the flipped fields in the d2 dataset. Examples are shown in the Figure~\ref{fig:flipped_fields}}
\label{tab:abliation_results_drifted}
\centering
\begin{tabular}{c | c c c | c}
    \hline
    \hline
     Different Features & Clean  & Drifted(flipped)  & Outliers & All \\
     \hline
     Spatial & 100 & 71.7 & 90 & 85.2 \\
     \hline
     Spatial+Aspect & 100 & 100 & 100 & 100 \\
    \hline
    \hline
\end{tabular}
\end{table}
\subsection{Ablation Study}\label{sec:ablation_study}
\begin{figure}[t]
	\centering
	\includegraphics[width = 1.0\linewidth]{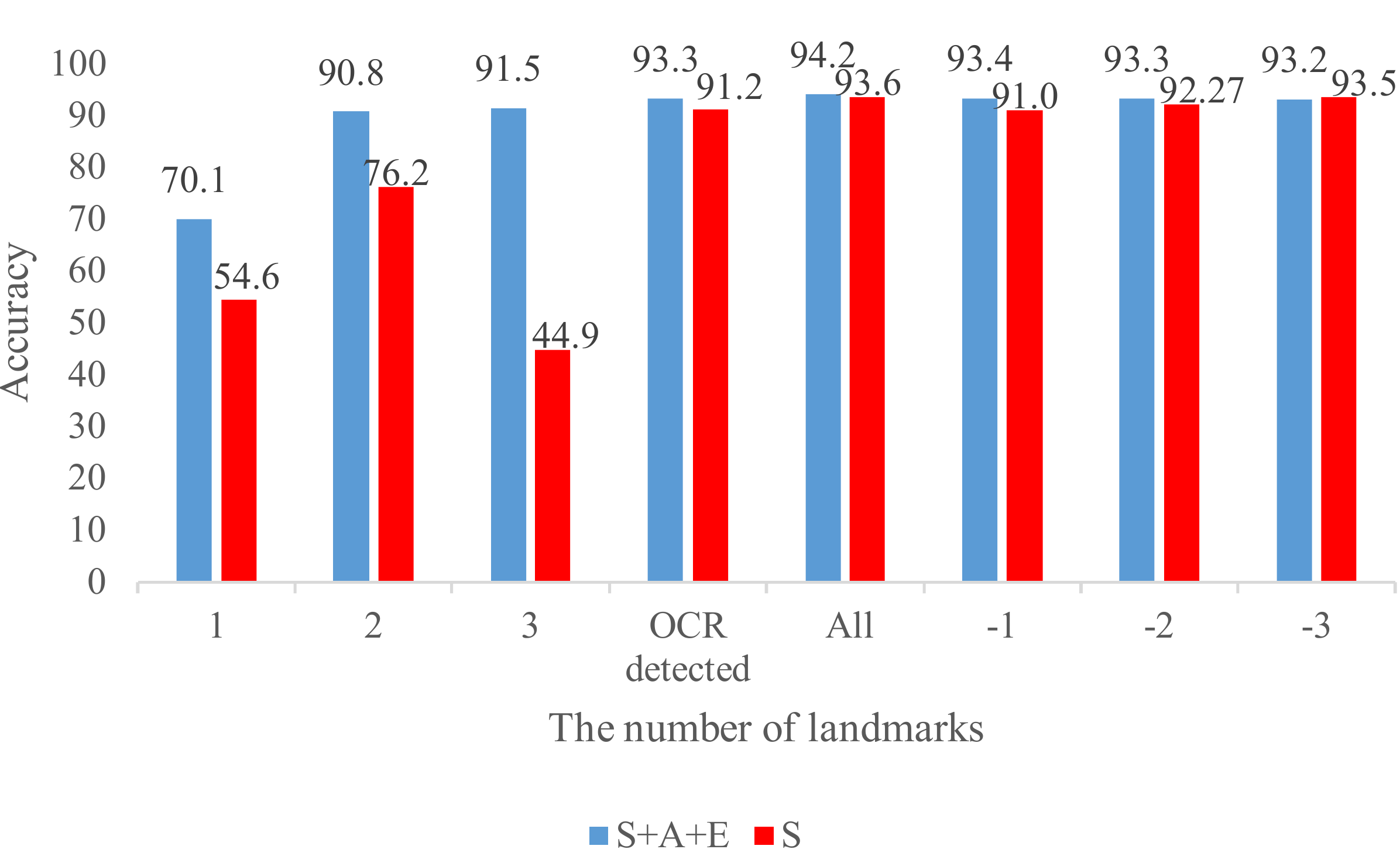}\\
	\bcaption{\label{fig:landmarkImpact}Labeling accuracy versus the number of landmarks. ``S'', ``A'', and ``E'' means spatial, aspect, and edge features.}
\end{figure}

We conduct ablation studies on the d0 and d2 dataset to examine the importance of spatial features, aspect features, textual features, edge features, the number of landmarks, and the ranking loss. \textcolor{black}{The d0 dataset contains 5 testing styles of documents. The d2 dataset contains 1 testing style. Each style has a separate support documents and a number of query documents. We use the d2 dataset to test whether different features can help handle the challenge of drifted fields and outliers. We present the results in Table~\ref{tab:abliation_results_drifted}. We use the d0 dataset to test the impact of different features across different styles. The results of training with different features are presented in Table~\ref{tab:abliation_results}. We report the influence of landmarks on the performance of our model in Fig. \ref{fig:landmarkImpact}. We also study the influence of ranking loss in Table~\ref{tab:ranking_loss_impact}.} 

\begin{table}[t]
\bcaption{Test the impact of different features using different styles of documents in the d0 dataset.  Document name alias: ``SC'': SiChuan province, ``BJ'': BeiJing, ``AH'': AnHui province, ``JS'': JiangSu province, ``CQ'': CongQing province. ``AVG'' indicates average accuracy.}
\label{tab:abliation_results}
\centering
\begin{tabular}{c | c c c c c | c}
    \hline
    \hline
     Different Features & SC  & BJ  & AH  & JS  & CQ  & AVG\\
     \hline
     Spatial & 98.8 & 91.2 & 81.8 & 92.8 & 99.1 & 93.6\\
     \hline
    Aspect & 0 & 0 & 0 & 0 & 0 & 0\\
    \hline
    Text & 10.2 & 10.4 & 14.7 & 7.2 & 12.3 & 10.6\\
     \hline
     Edge & 66.2 & 87.9 & 53.1 & 98.8 & 88.0 & 78.6\\
     \hline
     Spatial+Aspect & 98.8 & 85.3 & 94.4 & 94.0 & 97.2 & 93.2\\
     \hline
      Spatial+Edge & 97.2 & 87.9 & 95.8 & 91.6 & 99.1 & 93.2\\
     \hline
      Aspect+Edge & 91.6 & 82.2 & 74.1 & 98.2 & 90.0 & 87.1\\
     \hline
     Spatial+Aspect+Edge & 98.2 & 88.5 & 97.2 & 94.0 & 99.1 & 94.2 \\
     \hline
     Spa+Aspe+Text+Edge & 96.9 & 93.3 & 95.8 & 94.6 & 99.1 & \textbf{95.1} \\
    \hline
    \hline
\end{tabular}
\end{table}

\subsubsection{Different Features} \textcolor{black}{As shown in Figure~\ref{fig:flipped_fields} and Table~\ref{tab:abliation_results_drifted}, some fields are not possible to distinguish apart solely based on the spatial features. For example, in (a1) and (b1) of Figure~\ref{fig:flipped_fields}, the ``name$\_$chinese" field lies above the ``name$\_$english" field. In the contrary, their position exchanged in (a2) and (b2) of Figure~\ref{fig:flipped_fields}. According to the results in (a1) and (a2) of Figure~\ref{fig:flipped_fields}, if our model only uses spatial features, it maps the ``name$\_$chinese" field to the ``name$\_$english" field because they stay in the same position. However, two fields have very different width and height. Therefore, when our model incorporates the aspect of two fields to calculate their similarity scores, it can find the correct mapping between fields in (b1) and (b2) of Figure~\ref{fig:flipped_fields}. Table~\ref{tab:abliation_results_drifted} shows that our model can achieve 100 percent accuracy in the d2 dataset if it also uses the aspect features.}

\begin{figure}[t]
	\centering
	\includegraphics[width = 1.0\linewidth]{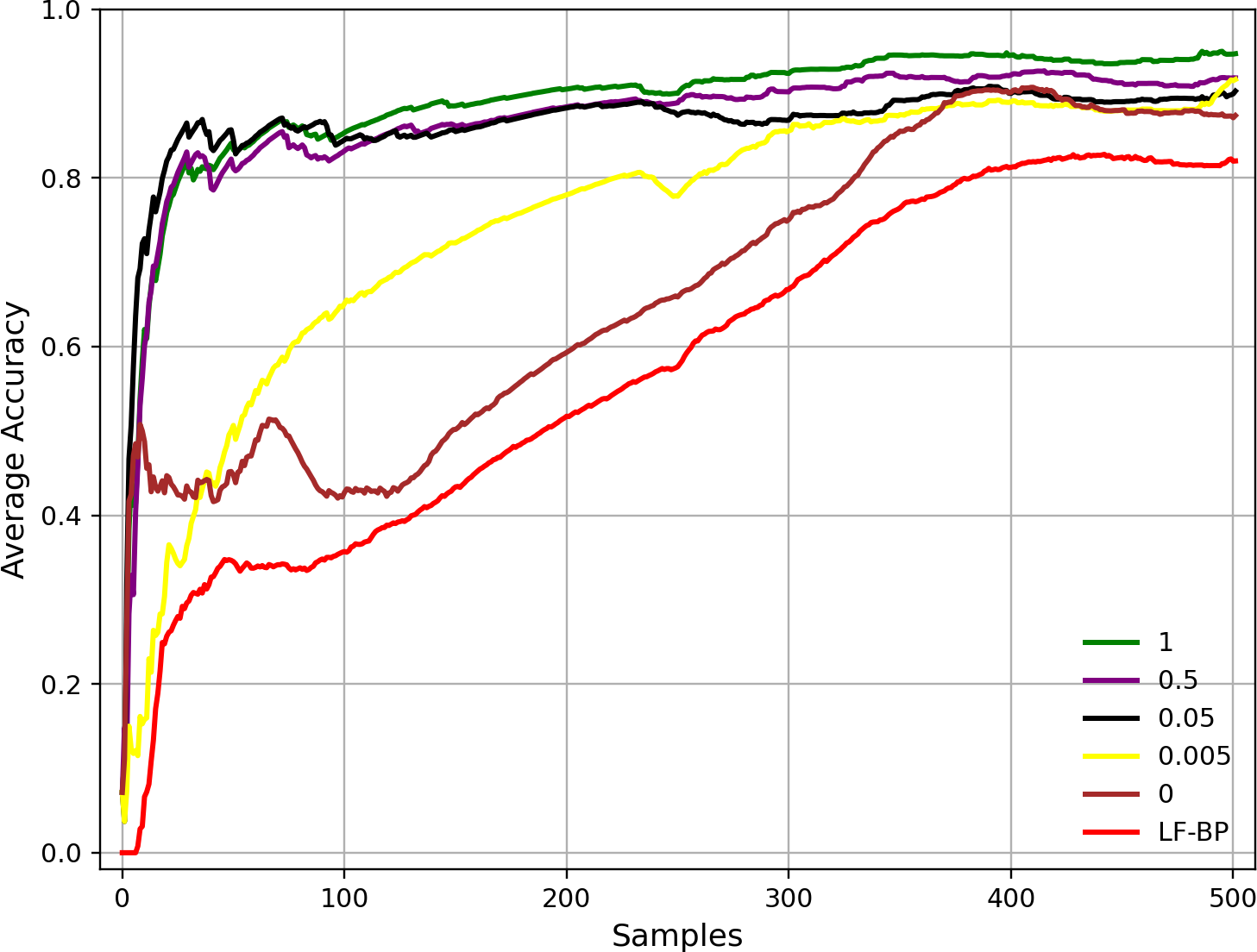}
	\caption{Accuracy of our models (ZAC-GM) that are trained with different ranking loss weight. We also include training accuracy of LF-BP model in this figure. Better viewed in color.}
	\label{fig:ranking_loss_ablation}
\end{figure}

\textcolor{black}{This inspired us to design additional MLP modules to incorporate more diverse features, and test the benefits of this practise across different datasets in Table~\ref{tab:abliation_results}.} First, we use only the spatial feature to train our model and test its performance. The first line of Table~ \ref{tab:abliation_results} indicates that our model achieves good performance solely based on the spatial features on most types of documents except for the ``AH'' type. The second line shows that our model fails if it only consumes the aspect features. This is not surprising because we find that many fields have a very similar shape and only a few of them have vast bounding boxes in the d0 dataset. Therefore, this feature alone can not distinguish the fields well. Similarly, our model will fail if it uses the textual features alone as we can see in the third line of Table~\ref{tab:abliation_results}. By checking the text of many fields, we found that many fields are hard to be distinguished by content, such as pure digits. We also find that our model will fail when the textual features are combined with other features except when all the features are used. Therefore, we do not report these failed results in Table~\ref{tab:abliation_results} to save the space. The forth line of Table~\ref{tab:abliation_results} tells that our model cannot perform well if we only use the edge features. By investigating the edges set of each document, we find that many edges in query documents cannot be found in their corresponding support documents. This is because the typology of the graph constructed for a query document can be very different from the one in the support document.

Second, we evaluate the model performance by combining two types of features to train our model. If we use ``Spatial+Aspect'' features (spatial combined with aspect features), the accuracy of our model on ``AH'' and ``JS'' type increases, while its accuracy on ``BJ'' and ``CQ'' decreases. If we use ``Spatial+Edge'' features (spatial combined with edge features), our model can perform better on the ``AH'' type, and worse on the ``SC'', ``BJ'', and ``JS'' type. Not surprisingly, if we use ``Aspect+Edge'' features (aspect combined with edge features), the performance of our model will only increase on the ``JS'' type, while decreases on the rest types. 
When we use ``Spatial+Aspect+Edge'' features, our model achieves the best accuracy on ``AH'' document with increases in accuracy from 14.7 to 97.2. Lastly, if we use all possible features (``Spatial+Aspect+Text+Edge''), the accuracy of our model on all types outperform 90\%. This is surprising because the textual features tends to decrease the performance of our model. Therefore, we believe the proposed four features are complementary to each other, and it is necessary to use all of them if possible.

\begin{table}[t]
    \centering
    \bcaption{Impact of Ranking Loss on different solvers.}
    \label{tab:ranking_loss_impact}
    \centering
    \begin{tabular}{c|c|c|c|c|c|c}
    \hline
    \hline
      & Solvers & SC & BJ & AH & JS & CQ\\ 
    \hline
    \multirow{2}{*}{Ranking loss (0)} & DD-ILP & 95.7 & 96.6 & 92.3 & 94.9 & 92.2 \\
    \cline{2-7}
     & ZAC-GM & 98.2 & 98.3 & 94.4& 92.5& 97.3\\
    \hline
    \multirow{2}{*}{Ranking loss (1)} & DD-ILP & 97.1 & 98.4 & 94.4 & 96.1 & 97.2\\
    \cline{2-7}
     & ZAC-GM & 98.8 & 99.2 & 98.8 & 99.8 & 99.1 \\
    \hline
    \hline
    \end{tabular}
\end{table}

\begin{table*}[t]
    \centering
    \caption{Different permutation matrix corresponds to different total affinity score.}
    \label{tab:different_total_affinity_score}
    \centering
    \begin{tabular}{c|c|c|c|c|c|c|c}
    \hline
    \hline
     \multicolumn{2}{c|}{\diagbox{(a1) in \\ Figure~\ref{fig:case_study_1}.}{Total:11.471 /\\ None}} & 
     \multicolumn{2}{c|}{\diagbox{(b1) in \\ Figure~\ref{fig:case_study_1}.}{Total:10.604 /\\10.481}} & 
     \multicolumn{2}{c|}{\diagbox{(c1) in \\ Figure~\ref{fig:case_study_1}.}{Total:9.782 /\\10.423}} & 
     \multicolumn{2}{c}{\diagbox{(d1) in \\ Figure~\ref{fig:case_study_1}.}{Total:11.257 /\\ None}}\\ 
    \hline
    Coordinates in $\hat{P}$ & Score & Coordinates in $\hat{P}$ & Score & Coordinates in $\hat{P}$ & Score & Coordinates in $\hat{P}$ & Score \\
    % pairs & \multirow{2}{*}{score} & pairs & \multirow{2}{*}{score} & pairs & \multirow{2}{*}{score} & pairs & \multirow{2}{*}{score}\\
    % (S,Q) &  & (S,Q) &  & (S,Q) &  & (S,Q) &  \\
    \hline
    \textcolor{black}{(S$\_$1,  Q$\_$7)} & \textcolor{black}{1.000} & (S$\_$1,  Q$\_$7) & 1.000 & (S$\_$1,  Q$\_$7) & 1.000 & (S$\_$1,  Q$\_$7) & 1.000\\
    \hline
    \textcolor{black}{(S$\_$1,  Q$\_$10)} & \textcolor{black}{0.980} & (S$\_$2,  Q$\_$2) & 0.959  & (S$\_$2,  Q$\_$2) & 0.959  & (S$\_$2,  Q$\_$10) & 0.937 \\
    \hline
    \textcolor{black}{(S$\_$1,  Q$\_$2)} & \textcolor{black}{0.964} & (S$\_$3,  Q$\_$11) & 0.937  & (S$\_$3,  Q$\_$11) & 0.937  & (S$\_$3,  Q$\_$2) & 0.910 \\
    \hline
    (S$\_$2,  Q$\_$11) & 0.938  & (S$\_$4,  Q$\_$3) & 0.879  & (S$\_$4,  Q$\_$3) & 0.879  & (S$\_$4,  Q$\_$11) & 0.902 \\
    \hline
    (S$\_$3,  Q$\_$3) & 0.880  & (S$\_$5,  Q$\_$8) & 0.863  & (S$\_$5,  Q$\_$8) & 0.863  & (S$\_$5,  Q$\_$3) & 0.861 \\
    \hline
    (S$\_$5,  Q$\_$8) & 0.863  & (S$\_$6,  Q$\_$1) & 0.842  & (S$\_$6,  Q$\_$1) & 0.842  & (S$\_$6,  Q$\_$8) & 0.845 \\
    \hline
    (S$\_$6,  Q$\_$1) & 0.842  & (S$\_$7,  Q$\_$6) & 0.832  & (S$\_$7,  Q$\_$6) & 0.832  & (S$\_$7,  Q$\_$1) & 0.823 \\
    \hline
    (S$\_$7,  Q$\_$6) & 0.832  & \textcolor{black}{(S$\_$8,  Q$\_$10)} & \textcolor{black}{0.122}  & (S$\_$8,  Q$\_$4) & 0.817  & (S$\_$8,  Q$\_$6) & 0.810 \\
    \hline
    \textcolor{black}{(S$\_$9,  Q$\_$4)} & \textcolor{black}{0.832}  & (S$\_$9,  Q$\_$4) & 0.832  & (S$\_$9,  Q$\_$9) & 0.821  & (S$\_$9,  Q$\_$4) & 0.832 \\
    \hline
    \textcolor{black}{(S$\_$9,  Q$\_$9)} & \textcolor{black}{0.821}  & (S$\_$10,  Q$\_$9) & 0.818  & (S$\_$10,  Q$\_$5) & 0.814  & (S$\_$10,  Q$\_$9) & 0.818 \\
    \hline
    (S$\_$11,  Q$\_$5) & 0.818  & (S$\_$11,  Q$\_$5) & 0.818  & (S$\_$11,  Q$\_$12) & 0.819  & (S$\_$11,  Q$\_$5) & 0.818 \\
    \hline
    (S$\_$12,  Q$\_$12) & 0.843  & (S$\_$12,  Q$\_$12) & 0.843  & (S$\_$12,  Q$\_$13) & 0.839 & (S$\_$12,  Q$\_$12) & 0.843 \\
    \hline
    (S$\_$13,  Q$\_$13) & 0.859  & (S$\_$13,  Q$\_$13) & 0.859  & \textcolor{black}{(S$\_$13,  Q$\_$10)} & \textcolor{black}{$-$0.641}  & (S$\_$13,  Q$\_$13) & 0.859 \\
    \hline
    \hline
    \end{tabular}
\end{table*}

\begin{figure}[!hbt]
	\centering
	\includegraphics[width = 0.8\linewidth]{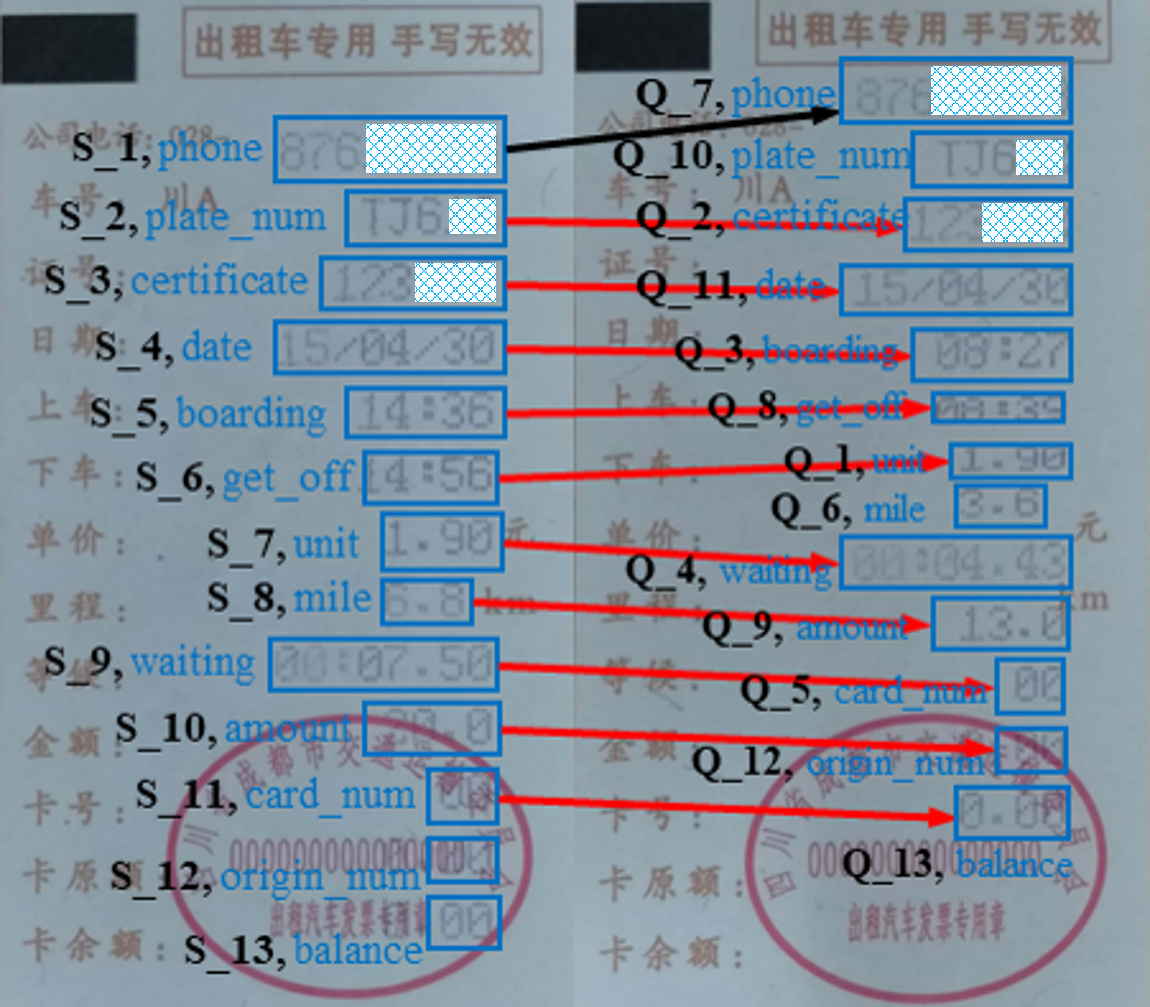}
	\caption{The prediction of LF-BP model on the drifted fields in d0 dataset.}
	\label{fig:ali_demo_mistake_in_d0}
\end{figure}

\subsubsection{Landmarks} We further evaluate the impacts of the number of landmarks on the accuracy of our model in Figure~\ref{fig:landmarkImpact}. In practice, text embedding (300 dimensions) may not be used to save computation resources, thus we only test the combination of ``Spatial+Aspect+Edge" (4 dimensions). The overall accuracy is good if we drop less than 3 landmarks (see -1, -2, -3 in the x-axis). When we use ``Spatial+Aspect+Edge'' features, the labeling accuracy grows as the number of landmarks increases. This is not true when we only use the spatial features. This also proves that these features are complementary to each other. 

\subsubsection{Ranking Loss} \textcolor{black}{We also evaluate the effectiveness of ranking loss by changing the weight of ranking loss. Figure~\ref{fig:ranking_loss_ablation} shows that our ranking loss can help accelerate the training process. When our model does not employ the ranking loss, our model can still outperform the LF-BP model. By increasing the weight of ranking loss, our model converged much more faster and the accuracy also increased. When the weight of ranking loss is too large, the performance of our model dropped. As shown in Table~\ref{tab:ranking_loss_impact}, when we apply the ranking loss to two solvers, the accuracy of them improved 1$\%$ across different testing styles in the d0 dataset.}

\subsection{Case Study}\label{subsec:case_study}
In this section, we presented a case study on the d0 dataset. The performance on this dataset can be found in Table~\ref{tab:over_all_results}. The accuracy of our model on this dataset is much higher than the LF-BP method~\cite{cheng2020one}, despite that both of them use the landmarks to generate spatial features. We select a pair of documents from the ``SC'' type as an example. As shown in Fig. \ref{fig:spatial_drift}, fields in the query document drift upwards when compared with fields in the support document. The LF-BP method successfully predicted fields with "receipt-code", "receipt-no", "phone-a", and "phone-b" labels. It failed on the rest fields because they drift and aligned with the wrong landmarks. Therefore, the LF-BP would mismatch their labels according to the landmarks. Although the Belief Propagation -`(BP) step in the LP-BP model may alleviate the spatial drift problem, both our re-implementation and 
their online demonstration $\footnote{https://ocr.data.aliyun.com/experience\#/?first\_tab=general}$ failed on this example.

\begin{figure*}[!hbt]
	\centering
	\includegraphics[width = 0.85\linewidth]{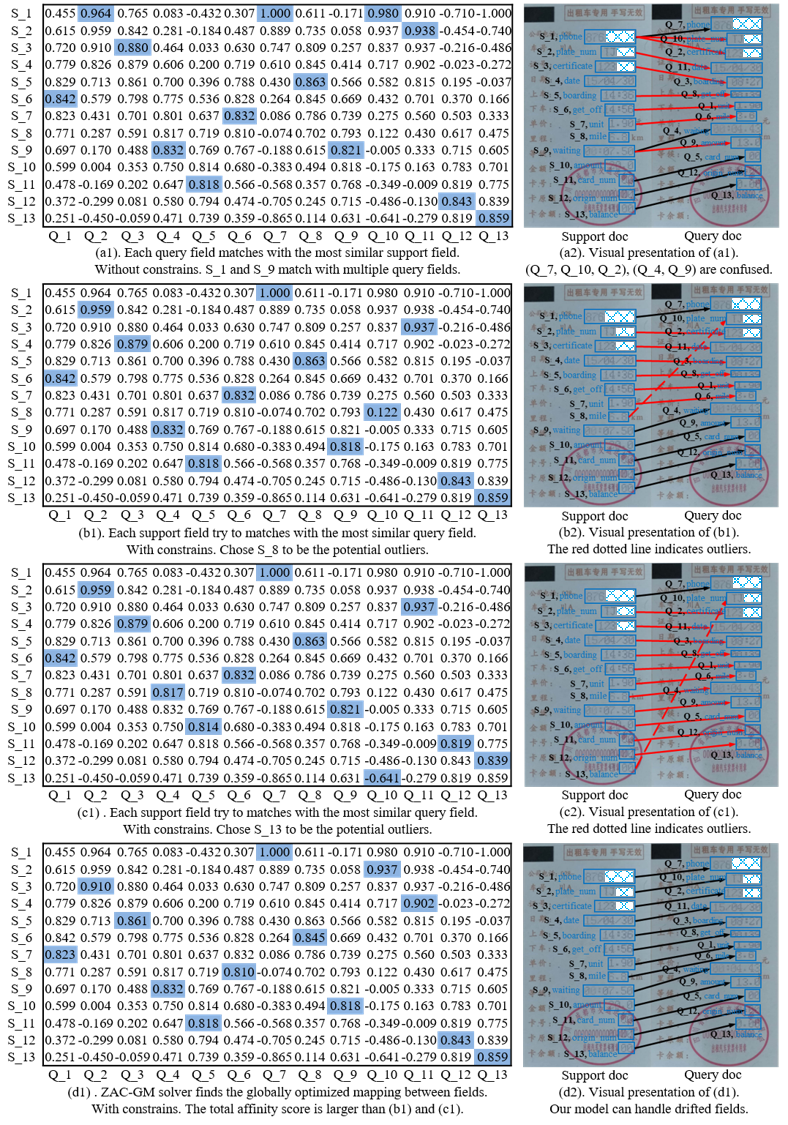}
	\caption{Different permutation matrix corresponds to different total affinity score.}
	\label{fig:case_study_1}
\end{figure*}

To dive into the details of the inference phase of our model, we visualized the vertex affinity matrix of our model in Fig. \ref{fig:case_study_1}. The fields in the query document are ordered in the same vertical sequence as shown in Fig.~\ref{fig:spatial_drift}. We deliberately disrupt the order of fields for the support document to add difficulty to the model. Each row of the affinity matrix in Fig.~\ref{fig:case_study_1} indicates the similarities between the current query filed and all the fields in the support document. Since the direct output of vertex MLP is not normalized, we apply the min-max normalization to each row of this matrix, i.e., each element subtracts the minimum value of its row, and then divided by the maximum value of its row. We will choose all "1" elements if we apply the greedy strategy used in LP-BP to select the  possible label for each query field. However, this solution conflicts with the one-to-(at most)-one constraint in our model. Therefore, combinatorial solvers in our model would find a globally optimized labeling strategy such that the overall affinity summation over all chosen elements is maximized and never violate the constrain. This enables that our model is less sensitive to vertex shift such that we can handle the spatial drifted cases well. Our model picked the elements lying in the red line path as shown in Fig. \ref{fig:case_study_1}. 

\section{Conclusion}
In this work, we proposed to solve the text field labeling problem using graph matching. We designed a one-shot framework that combines the power of deep learning and combinatorial solvers. To the best of our knowledge, our framework is the first to generate globally optimized solutions. Our model could handle spatial drifted documents, and shows state-of-the-art performance on most testing datasets.

Other potential visual cues, such as text color, fonts, and background, will be explored in future work in addition to current textual, aspect, and spatial relationship features. Our method can be extended to support few-short learning by adding additional constraints such as cycle consistency, and we leave it for future work.

% or


% \appendices

% % you can choose not to have a title for an appendix
% % if you want by leaving the argument blank
% \section{}

% % use section* for acknowledgment
% \section*{Acknowledgment}


\ifCLASSOPTIONcaptionsoff
  \newpage
\fi



\bibliographystyle{IEEEtran}
\bibliography{My_bib}
\end{document}